\documentclass[lettersize, journal]{IEEEtran}

\usepackage[final]{changes}
\usepackage{changes}
\usepackage{amsmath, amsfonts}
\usepackage{algorithmic}
\usepackage{algorithm}
\usepackage{array}
\usepackage[caption=false, font=normalsize, labelfont=sf, textfont=sf]{subfig}
\usepackage{textcomp}
\usepackage{stfloats}
\usepackage{url}
\usepackage{verbatim}
\usepackage{graphicx}
\usepackage{cite}
\usepackage{amssymb}
\usepackage{graphics} 
\usepackage{epsfig}
\usepackage{booktabs}
\usepackage{multirow}
\usepackage[utf8]{inputenc}
\usepackage{url}
\usepackage{booktabs}
\usepackage{amssymb}
\usepackage{bbding}
\usepackage{pifont}
\usepackage{wasysym}
\usepackage{utfsym}
\usepackage{fontawesome}
\usepackage{booktabs}
\usepackage{multirow}
\usepackage{amsmath}
\usepackage{hyperref}
\usepackage{mathrsfs}
\usepackage[numbers, sort&compress]{natbib}
\usepackage{txfonts}
\begin{document}

\title{SCTransNet: Spatial-channel Cross Transformer Network for Infrared Small Target Detection}

\author{Shuai~Yuan~\IEEEmembership{Student~Member,~IEEE},~Hanlin~Qin~\IEEEmembership{Member,~IEEE},~Xiang~Yan~\IEEEmembership{Member,~IEEE},~Naveed~Akhtar~\IEEEmembership{Member,~IEEE},~Ajmal~Mian~\IEEEmembership{Senior Member,~IEEE}
% \authornote{*}
        % <-this % stops a space
% \thanks{This paper was produced by the IEEE Publication Technology Group. They are in Piscataway, NJ.}% <-this % stops a space. Manuscript received April 19, 2021; revised August 16, 2021.  
 % in part by the Ministry of Education Joint Foundation 8091B032233,
\thanks{This work was supported in part by the Shaanxi Province Key Research and Development Plan Project under Grant 2022JBGS2-09, in part by the 111 Project under Grant B17035, in part by the Shaanxi Province Science and Technology Plan Project under Grant 2023KXJ-170, in part by the Xian City Science and Technology Plan Project under Grant 21JBGSZ-QCY9-0004, Grant 23ZDCYJSGG0011-2023, Grant 22JBGS-QCY4-0006, and Grant 23GBGS0001, in part by the Aeronautical Science Foundation of China under Grant 20230024081027, in part by the Natural Science Foundation Explore of Zhejiang province under Grant LTGG24F010001, in part by the Natural Science Foundation of Ningbo under Grant 2022J185, in part by the China scholarship council 202306960052, in part by the Technology Area Foundation of China 2021-JJ-1244, 2021-JJ-0471, 2023-JJ-0148, and part by the Xidian Graduate Student Innovation fund under Grant YJSJ23010. \textit{(Corresponding authors:~Hanlin Qin, Xiang Yan.)}

Shuai~Yuan, Hanlin~Qin, and Xiang~Yan are with the School of Optoelectronic Engineering, Xidian University, Xi'an 710071, China. (email: yuansy@stu.xidian.edu.cn; hlqin@mail.xidian.edu.cn; xyan@xidian.edu.cn)

Naveed Akhtar is with the School of Computing and Information Systems, Faculty of Engineering and IT, The University of Melbourne, Parkville VIC 3052, Australia (email: naveed.akhtar1@unimelb.edu.au).

Ajmal Mian is with the Department of Computer Science and Software Engineering, The University of Western Australia, Perth, 6009, Australia (email: ajmal.mian@uwa.edu.au).
}
}

\markboth{Journal of \LaTeX\ Class Files,~Vol.~14, No.~8, August~2021}%
{Shell \MakeLowercase{et al.}: A Sample Article Using IEEEtran.cls for IEEE Journals}

\maketitle
\begin{abstract}
\textcolor{blue}{This is the pre-acceptance version, to read the final version please go to IEEE TRANSACTION ON GEOSCIENCE AND REMOTE SENSING on IEEE Xplore.}
Infrared small target detection (IRSTD) has recently benefitted greatly from U-shaped neural models. 
However, largely overlooking effective global information modeling, existing techniques struggle when the target has high similarities with the background.
We present a \emph{S}patial-channel \emph{C}ross \emph{T}ransformer \emph{Net}work~(SCTransNet) that leverages spatial-channel cross transformer blocks~(SCTBs) on top of long-range skip connections to address the aforementioned challenge.
In the proposed SCTBs, the outputs of all encoders are interacted with cross transformer to generate mixed features, which are redistributed to all decoders to effectively reinforce semantic differences between the target and clutter at full levels.
Specifically, SCTB contains the following two key elements:
(a) spatial-embedded single-head channel-cross attention~(SSCA) for exchanging local spatial features and full-level global channel information to eliminate ambiguity among the encoders and facilitate high-level semantic associations of the images, 
and (b) a complementary feed-forward network~(CFN) for enhancing the feature discriminability via a multi-scale strategy and cross-spatial-channel information interaction to promote beneficial information transfer.
Our SCTransNet effectively encodes the semantic differences between targets and backgrounds to boost its internal representation for detecting small infrared targets accurately.
Extensive experiments on three public datasets, NUDT-SIRST, NUAA-SIRST, and IRSTD-1K, demonstrate that the proposed SCTransNet outperforms existing IRSTD methods. Our code will be made public at 
\url{https://github.com/xdFai/SCTransNet}.

\end{abstract}

\section{Introduction}
\label{sec:IN}
\IEEEPARstart{I}{nfrared} small target detection (IRSTD) plays an important role in traffic monitoring~\cite{54}, maritime rescue~\cite{41}, and target warning~\cite{55}, where separating small targets in complex scene backgrounds is required.
The challenges emerging from the dynamic nature of scenes have attracted considerable research attention in single-frame IRSTD~\cite{64}. Early methods in this direction employed image filtering~\cite{1},~\cite{2}, human visual system (HVS)~\cite{3},~\cite{4}, and \replaced{low-rank}{low rank} approximation~\cite{5},~\cite{6} techniques while relying on complex handcrafted feature designs, empirical observations, and model parameter fine-tuning.
However, suffering from the absence of a reliable high-level understanding of the holistic scene, these methods exhibit poor robustness.

Recently, learning-based methods have become more popular due to their strong data-driven feature mining abilities~\cite{59}. 
To capture the target's outlines and mitigate performance degradation caused by its small size, these methods approach the IRSTD problem as a semantic segmentation task instead of a traditional object detection issue.
Unlike general object segmentation in autonomous driving~\cite{38}, imaging mechanism of the IR detection systems in remote sensing applications~\cite{56} leads to small targets in images exhibiting the following characteristics. 
\textbf{1) Dim and small:} Due to remote imaging, IR targets are small and usually exhibit a low signal-to-clutter ratio, 
making them susceptible to immersion in heavy noise and background clutter.
\textbf{2) Characterless:} Thermal images lack color and texture information in targets, and imprecise camera focus can cause target blurring. These factors pose peculiar challenges in designing feature extraction techniques for IRSTD.
\textbf{3) Uncertain shapes:} The scales and shapes of IR targets vary significantly across different scenes, which makes the problem of detection considerably challenging. 

\begin{figure*}[t]
    \centering
    \includegraphics[width=17.5cm]{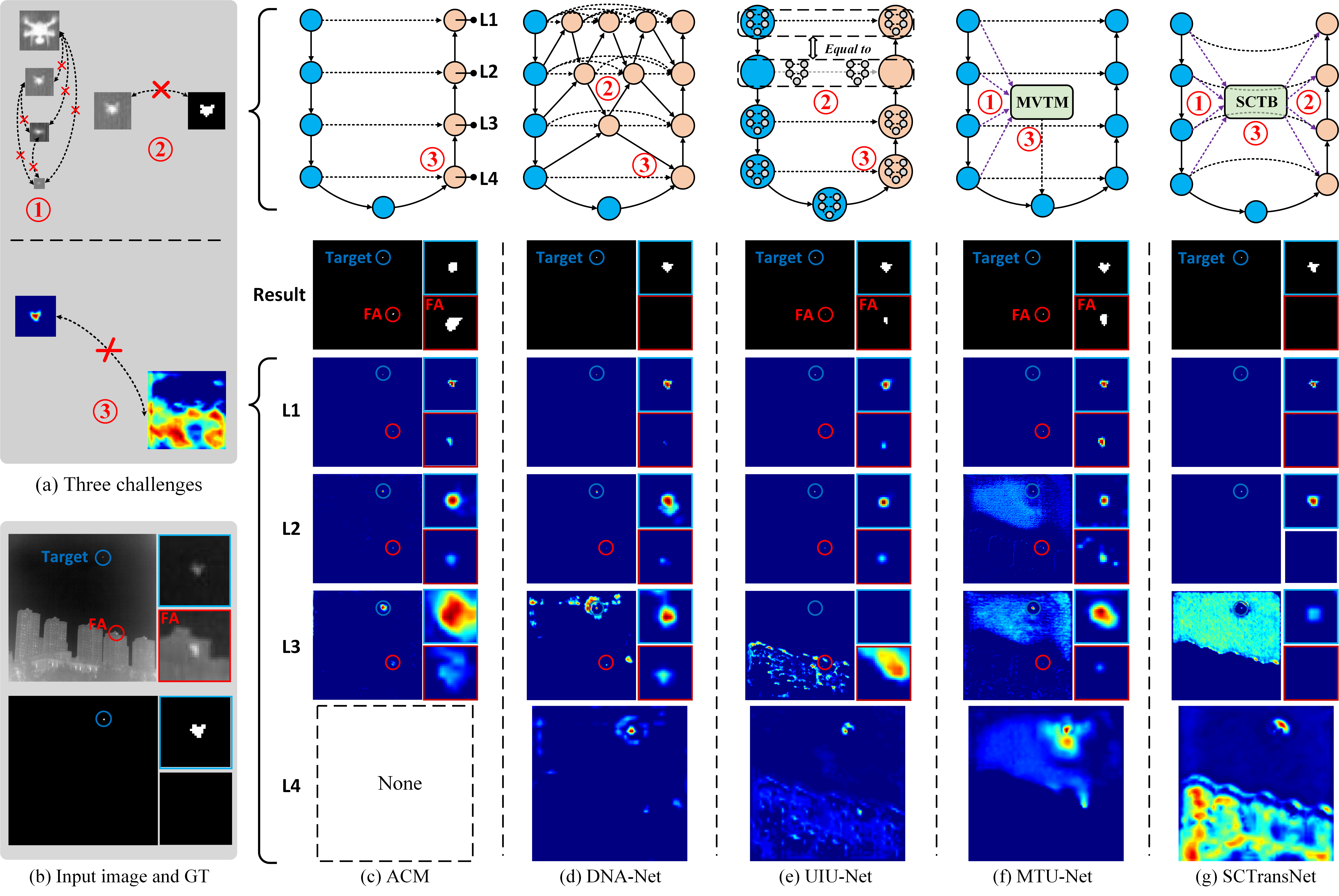}
    \vspace{-1mm}
    \caption{The framework and visualization maps of representative IRSTD methods, with each method's frame labeled according to the specific challenge it addresses. The visualization maps show that the CNN-based approaches (ACM~\cite{7}, DNA-Net~\cite{8}, and UIU-Net~\cite{10}) focus on modeling the local information of the target and less on establishing the global semantic information of the image; Mixer CNN and transformer methods (MTU-Net~\cite{11} and our SCTransNet) pay more attention to the background global information and target semantics. Only our method meticulously models buildings and the sky separately in the high-level semantic map, effectively distinguishing the target from the background, reducing false alarms. 
    }
    \vspace{-3mm}
    \label{fig1}
\end{figure*}

To identify small IR targets in complex backgrounds, numerous learning-based methods have been proposed, among which 
\replaced{neural networks with U-shaped architectures}{U-shaped architectures of neural networks} have gained prominence. 
\replaced{Benefiting from these frameworks of encoders, decoders, and long-range skip connections,}{These networks consist of encoders, decoders, and long-range skip connections.} 
asymmetric contextual modulation (ACM) network~\cite{7} initially demonstrated the effectiveness of cross-layer feature fusion for retaining IR target features. This is achieved through bidirectional aggregation of high-level semantic information and low-level details using asymmetric top-down and bottom-up structures.
Subsequently, feature fusion strategies have been widely adopted in IRSTD task~\cite{29},~\cite{30},~\cite{53},~\cite{60}.
A few recent methods facilitate the transfer of beneficial features to the decoder component by improving the skip connections~\cite{9},~\cite{36}.
Inspired by the nested structure~\cite{57}, 
% DNA-Net~\cite{8} designed a dense nested interactive module to achieve progressive interaction between high- and low-level features and adaptive feature enhancement.
DNA-Net~\cite{8} developed a densely nested interactive module to facilitate gradual interaction between high- and low-level features and adaptively enhance features.
Moreover, there are also approaches that focus on developing more effective encoders and decoders~\cite{51},~\cite{52}. For instance, UIU-Net~\cite{10} embeds smaller U-Nets in the U-Net to learn the local contrast information of the target and perform interactive-cross attention (IC-A) for feature fusion.

Despite achieving satisfactory results, the aforementioned CNN-based approaches lack the ability to encode comprehensive attributes of the target, missing their discriminative features.
To address that, MTU-Net~\cite{11} employs a multilevel Vision Transformer (ViT)-CNN hybrid encoder to exploit the spatial correlation among all encoded features for contextual information aggregation.
However, a simple spatial ViT-CNN hybrid module is insufficient for understanding the global semantics of images, which makes high false alarms.
To further dissect the issue, we illustrate the frameworks of ACM~\cite{7}, DNA-Net~\cite{8}, UIU-Net~\cite{10}, and MTU-Net~\cite{11} separately, along with visualizations of the attention maps from different decoder levels in Fig.~\ref{fig1}(c)-(f). 
\added{Given the input image in Fig.~\ref{fig1}(b),}
we observe that false alarms occur when existing models direct their attention to localized regions of background clutters in high-level features. In other words, false alarms are often caused by discontinuity modeling of backgrounds in the deeper layers.
We identify this problem to the following three main reasons:

% Existing methods give high false alarms due to the following three main reasons:

\textbf{1) Semantic interaction across feature levels is not established well}. 
%IR small targets exhibit limited size and features.
% IR small targets exhibit limited features due to their size. 
% $\textcircled{1}$
\added{As shown in Fig.~\ref{fig1}(a)\ding{172}}, IR small targets exhibit limited features owing to their diminutive size.
Multiple downsampling processes inevitably result in the loss of %image 
spatial information. This considerably affects the level-to-level feature interactions in the network, eventually leading to poor comprehensive global semantic information encoding.

\textbf{2) Feature enhancement fails to bridge the information gap between encoders and decoders}.
\added{As shown in Fig.~\ref{fig1}(a)\ding{173}}, there exists a semantic gap between the output features of encoders and the input features of the decoders. Simple skip connections and dense nested modules are insufficient to enhance the advantageous responses of the features to the decoder, thereby making it challenging to establish a mapping relationship from the IR image to the segmentation space.

\textbf{3) Inaccurate long-range contextual perception of targets and backgrounds in deeper layers}.
IR small targets can be highly similar to the scene background. \added{As shown in Fig.~\ref{fig1}(a)\ding{174}}, a powerful detector not only has to sense the local saliency of the target but also needs to model the continuity of the background. Convolutional Neural Networks (CNNs) and vanilla ViTs are not fully equipped to achieve this.

\added{Inspired by the success of channel-wise cross fusion transformer in image segmentation~\cite{23},~\cite{24},~\cite{25} and local spatial embedding in image restoration~\cite{22},~\cite{67},~\cite{48},}
\replaced{we propose a spatial-channel cross transformer network (SCTransNet) for IRSTD to address the above challenges, }{To address the above problems, we propose a spatial-channel cross transformer network (SCTransNet) for IRSTD,}
aiming to distinguish the small targets and background clutters in deeper layers. 
As illustrated in Fig.~\ref{fig1}(g), our framework adds multiple spatial-channel cross transformer blocks (SCTB) (Sec.~\ref{sec: SCTB}) on the original skip connections to establish an explicit association with all encoders and decoders.
Specifically, SCTB consists of two components: Spatial-embedded single-head channel-cross attention (SSCA) (Sec.~\ref{labSSCA}) and complementary feed-forward network (CFN) (Sec.~\ref{labCFN}).

The SSCA applies channel cross-attention from the feature dimension at all levels to learn global information. Besides, depth-wise convolutions are used for local spatial context mixing before feature covariance computation.
This strategy provides two advantages: 
Firstly, it highlights the context of local space with a small computational overhead using the convolution's local connectivity, thereby increasing the saliency of IR small targets.
Secondly, it makes sure that contextualized global relationships among full-level feature pixels are implicitly captured during the attention matrix computation, thereby reinforcing the continuity of the background.

After the SSCA completes the cross-level information interaction, CFN performs feature enhancement at every level in two complementary stages.
Initially, it utilizes multi-scale depth-wise convolutions to enhance target neighborhood space response and pixel-wise aggregates the cross-channel nonlinear information.
Subsequently, it estimates total spatial information on a channel-by-channel basis using global average pooling and creates local cross-channel interactions between distinct semantic patterns as an attention map.
The above strategy has two advantages. 
(1) Multi-scale spatial modeling can emphasize semantic differences between the target and background.
(2) Establishing the complementary correlation of the local space global channel (LSGC) and the global space local channel (GSLC) can facilitate the interface between infrared images and semantic maps.

Benefiting from the above structure (Fig.~\ref{fig1}(g)), our SCTransNet can perceive the image semantics better than other methods leading to reduced false alarms. 
Our main contributions are as follows:
\begin{itemize}
    \item We propose SCTransNet, which leverages multiple spatial-channel cross transformer blocks (SCTB) connecting all encoders and decoders to predict the context of targets and backgrounds in the deeper network layers. 
    \item We propose a spatial-embedded single-head channel-cross attention (SSCA) module to foster semantic interactions across all feature levels and learn the long-range context correlation of the image.
    \item We devise a novel complementary feed-forward network (CFN) by crossing spatial-channel information to enhance the semantic difference between the target and background, bridging the semantic gap between encoders and decoders. 
\end{itemize}

\vspace{-2mm}
\section{RELATED WORK}
\vspace{-1mm}
\label{Sec2}
We first briefly review the CNN- and transformer-based techniques in IRSTD. Following that, we discuss the application of channel-wise cross transformer in image processing.

\subsection{CNN-based IRSTD methods}
\label{Sec2-A}
Owing to the local saliency of IR small targets coinciding with the local connectivity of convolution neural networks (CNNs), CNNs have demonstrated remarkable performance in the IRSTD task.
To effectively preserve the semantic patterns of small targets, diverse feature fusion strategies have been proposed.
One common strategy is cross-layer feature fusion~\cite{31},~\cite{40},~\cite{61}, which can address the loss of target information when fusing the encoded and decoded features.
Additionally, densely nested interactive feature fusion~\cite{8},~\cite{37} is used to repetitively fuse and enhance the features of different levels, maintaining the information of IR small targets in the deeper layers.
Considering variations in target scales, multi-scale feature fusion~\cite{34},~\cite{35} has been proposed to enhance the low-resolution feature maps.
Besides feature fusion, incorporating prior information about the target into CNNs is also an effective strategy. For instance, Sun et al.~\cite{12} exploited the small-target gray gradient change property using a receptive-field and direction-induced attention network (RDIAN), which solves the imbalance between the target and background classes.
Zhang et al.~\cite{13} used Taylor's finite difference for complex edge feature extraction of a target to enhance the target and background gray scale difference.

Although satisfactory results are achieved by CNN-based techniques, the inherent inductive bias of CNNs makes it difficult to unambiguously establish long-range contextual information for the IRSTD task.
Unlike the aforementioned methods, we incorporate transformer blocks into the backbone of CNNs as a core unit to capture non-local information for the entire image.

\begin{figure*}[t]
    \centering
    \includegraphics[width=17.2cm]{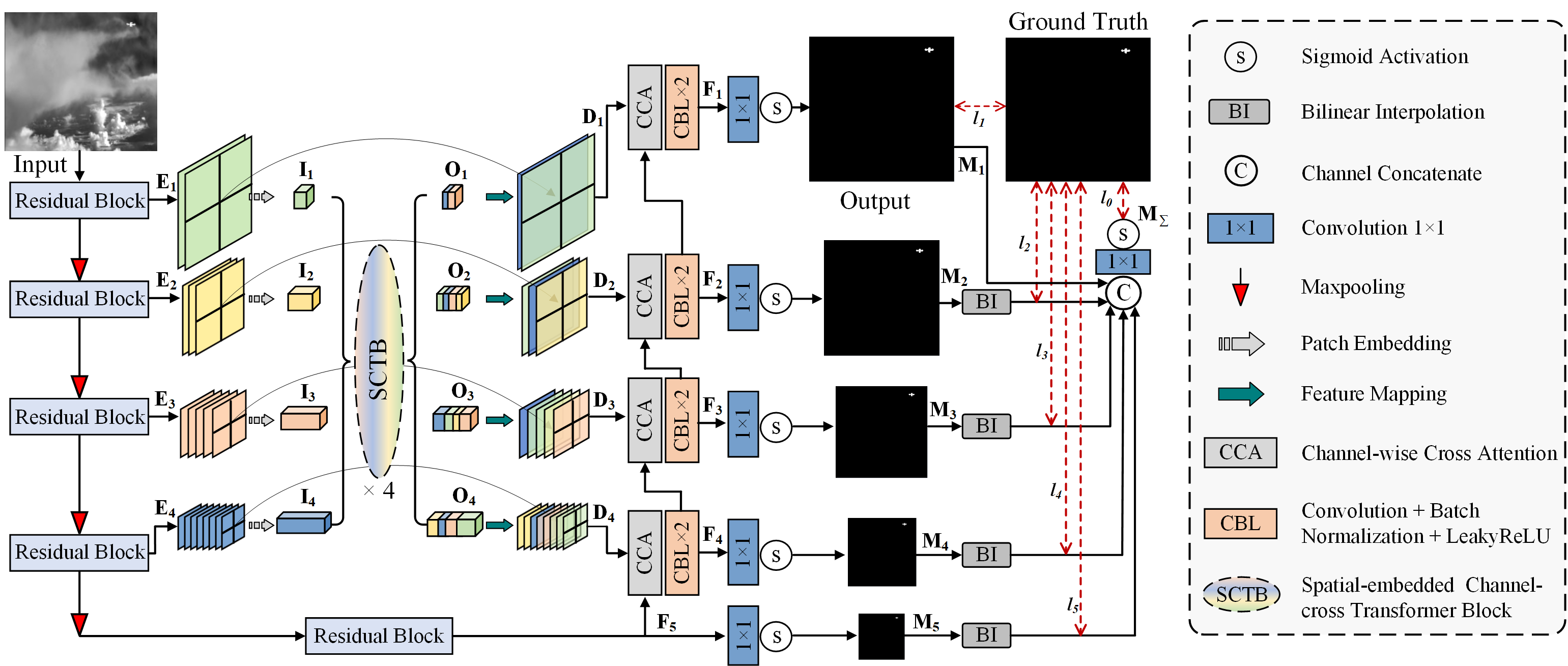}
    \vspace{-1mm}
    \caption{\added{Overview of the proposed SCTransNet for infrared small object detection. Our SCTransNet adopts a U-shaped structure and adds four spatial-channel cross transformer blocks (SCTB) on the long-range skip connections, and the multi-scale deeply supervised fusion strategy is used to optimize our SCTransNet.}
    }
    \vspace{-1mm}
    \label{fig2}
\end{figure*}

\subsection{Transformer-based IRSTD methods}
Vision Transformer~(ViT)~\cite{14} decomposes an image/features into a series of patches and computes their correlation. This computational paradigm can stably establish long-distance dependence among different patches, \replaced{leading to}{which has led to} its widespread usage in IRSTD tasks for global image modeling~\cite{15},~\cite{16},~\cite{27}.
Inspired by TransUnet~\cite{19}, IRSTFormer~\cite{20} embedded the spatial transformer within multiple encoder stages in a U-Net. Motivated by Swin transformer~\cite{18}, FTC-Net~\cite{17} establishes a robust feature representation of the target using a two-branch structure combining the local feature extraction of CNNs and the global feature extraction capability of the Swin transformer.
Recently, Meng et al.~\cite{21} modeled the local gradient information of the target using central difference convolution and employed criss-cross multi-attention~\cite{65} to acquire contextual information.  
% at a low computational overhead
Note that, the above methods use spatial self-attention (SA) to calculate covariance-based attention maps, which have two problems: 1) The computational complexity is proportional to the square of the number of tokens, which limits the multiple nesting of the spatial transformer and its fine-grained representation of high-resolution images~\cite{22}.
2) The SA only constructs long-distance dependency for a single feature map, whereas it is more critical to establish contextual connections among all levels.

Different from previous works, we present the channel-wise cross transformer on the long-range skip connections for the first time in the IRSTD task.
This allows establishing cross-channel semantic patterns across all levels with an acceptable computational overhead.

\subsection{Channel-wise Cross Transformer on Image Processing}
Unlike spatial transformers, channel-wise transformers (CT)~\cite{22} treat each channel as a patch. Note that every channel is a unique semantic pattern, CT essentially establishes correlations between multiple semantic patterns.
Considering that not every skip connection is effective, 
\replaced{Wang et al.~\cite{23} proposed UCTransNet, utilizing channel-wise cross fusion transformer (CCT) to address the semantic difference for precise medical image segmentation.}{Wang et al.~\cite{23} proposed a channel-wise cross fusion transformer (CCT) to address the semantic difference for precise medical image segmentation.}
The CCT's powerful global semantic modeling capability facilitates its widespread application in tasks such as metal surface defect detection~\cite{25}, remote sensing image segmentation~\cite{26}, and building edge detection~\cite{24}. 
This inspires us to introduce this model to separate IR targets and backgrounds in the deeper layers effectively.
\replaced{However, IR small targets differ significantly from the usual large-size targets not only in size but also in terms of effective features and sample balance.}{However, the IR small targets differ significantly from the usual large-size targets not only in terms of target size, but also in effective features, and sample balance.}
The attention matrix computation, the positional encoding, and the pure channel modeling in vanilla CCT are harmful to the limited-pixel target detection.
Therefore, we propose a spatial-channel cross transformer block. Its launching point is leveraging the target's local spatial saliency and global background continuity to separate the target in the deep layers.

\vspace{-1mm}
\section{METHOD}
\vspace{-1mm}
This section elaborates on the proposed Spatial-channel Cross Transformer Network (SCTransNet) for infrared small target detection. We begin by presenting the overall structure of the proposed SCTransNet in Section~\ref{sec: PIP}. Then, we present the technical details of the spatial-channel cross transformer block~(SCTB) and its internal structure: Spatial-embedded single-head channel-cross attention~(SSCA) and the complementary feed-forward network~(CFN) in Section~\ref{sec: SCTB}.

\subsection{Overall pipeline}
\label{sec: PIP}
As shown in Fig.~\ref{fig2}, given an infrared image, SCTransNet initially employs four groups of \replaced{residual blocks}{ResBlocks} (RBs)~\cite{28} and max-pooling layers, to acquire high-level features ${\mathbf{{E}_{i}}} \in \mathbb{R}^{{C_{i}} \times {\frac{H}{i}} \times {\frac{W}{i}}}$, $(i = 1, 2, 3, 4)$. ${C_{i}}$ are the channel dimensions, in which ${C_{1}}$ = 32, ${C_{2}}$ = 64, ${C_{3}}$ = 128, ${C_{4}}$ = 256.
Next, we perform patch embedding on $\mathbf{{E}_{i}}$ using convolution with kernel size and stride size of $P$, $P/2$, $P/4$, and $P/8$ to obtain embedded layers ${\mathbf{{I}_{i}}} \in \mathbb{R}^{{C_{i}} \times {\frac{H}{16}} \times {\frac{W}{16}}}$, $(i = 1, 2, 3, 4)$ respectively.
These layers are then fed into the SCTB for full-level semantic feature blending and obtaining the output
${\mathbf{{O}_{i}}} \in \mathbb{R}^{{C_{i}} \times {\frac{H}{16}} \times {\frac{W}{16}}}$, $(i = 1, 2, 3, 4)$, which have the same size of ${\mathbf{{I}_{i}}}$. Details of SCTB are provided in the next Section. 
The ${\mathbf{{O}_{i}}}$ are recovered to the size of the original encoder processing using feature mapping (FM), which consists of bilinear interpolation, convolution, \replaced{batch normalization}{batchnorm}, and ReLU activation.  Meanwhile, we employ a residual connection to merge the features between the encoders and decoders. The  process described above can be expressed mathematically as 
\begin{equation}
\mathbf{O_{i}}={\mathbf{E_{i}}}+{\text{FM}_{i}}(\text{SCTB}(\mathbf{I_{1}}, \mathbf{I_{2}}, \mathbf{I_{3}}, \mathbf{I_{4}}))~(i = 1, 2, 3, 4).
\end{equation}
Finally, the Channel-wise Cross Attention (CCA)~\cite{23} is employed to fuse the high- and low-level features, followed by decoding using two CBL blocks.

To enhance the gradient propagation efficiency and feature representation, we utilize a multi-scale deeply supervised fusion strategy to optimize SCTransNet.
Specifically, a $1\times1$ convolution and sigmoid function are used for each decoder outputs $\mathbf{F_{i}}$, acquiring the saliency map $\mathbf{M_{i}}$ which is denoted as
\begin{equation}
% \label{1}
\mathbf{M_{i}}=\text{Sigmoid}({f_{1\times1}}(\mathbf{F_{i}}))~(i = 1, 2, 3, 4, 5).
\end{equation} 
Next, we upsample the low-resolution salient maps $\mathbf{M_{i}}~(i = 2, 3, 4, 5)$ to the original image size and fuse all the salient maps to obtain $\mathbf{M_{\sum}}$ as
\begin{equation}
\mathbf{M_{\sum}}=\text{Sigmoid}({f_{1\times1}}[\mathbf{M_{1}}, \mathcal{B}({\mathbf{M_{2}}), \mathcal{B}(\mathbf{M_{3}}), \mathcal{B}(\mathbf{M_{4}}), \mathcal{B}(\mathbf{M_{5}})}]),
\end{equation}
where $[\cdot]$ is the channel-wise concatenation, $\mathcal{B}$ denotes the bilinear interpolation.
Finally, we calculate the Binary Cross Entropy (BCE)~\cite{10} loss between the overall saliency maps and the ground truth (GT) $\textbf{Y}$ as below, and combine the losses.
\begin{align}
& {l_{1}}= {\mathcal{L}_{BCE}}(\mathbf{M_{1}}, \mathbf{Y}), \\
& {l_{i}}= {\mathcal{L}_{BCE}}(\mathcal{B}(\mathbf{M_{i}}), \mathbf{Y})~(i = 2, 3, 4, 5), \\
& {l_{\sum}}= {\mathcal{L}_{BCE}}(\mathbf{M_{\sum}}, \mathbf{Y}), \\
& L ={\lambda_{1}}{l_{1}}+{\lambda_{2}}{l_{2}}+{\lambda_{3}}{l_{3}}+{\lambda_{4}}{l_{4}}+{\lambda_{5}}{l_{5}}+{\lambda_{\sum}}{l_{\sum}},
\end{align}
in which ${\lambda_{i}~(i =1, 2, 3, 4, 5)}$ represents the weights corresponding to different loss functions. In this work, ${\lambda_{i}}$ and ${\lambda_{\sum}}$ are set to 1 empirically.

\begin{figure*}[t]
    \centering
    \includegraphics[width=16.8cm]{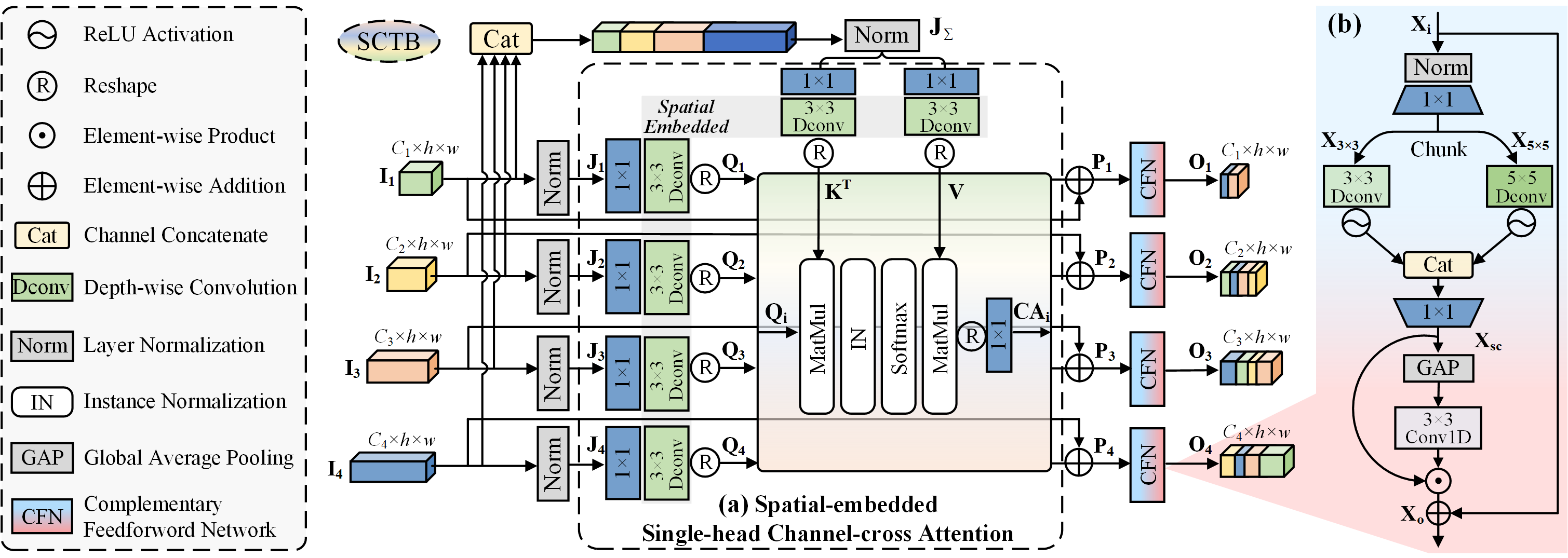}
    % %\vspace{-3mm}
    \caption{\added{The proposed spatial-channel cross transformer block (SCTB), which consists of spatial-embedded single-head channel-cross attention (SSCA) and complementary feed-forward network (CFN). (a) SSCA establishes image full-scale information association by means of different levels of semantic interaction. (b) CFN bridges the semantic gap between encoder and decoder through complementary feature enhancement.
    }
    }
    \label{fig3}
\end{figure*}

\subsection{Spatial-channel Cross Transformer Block}
\label{sec: SCTB}
Recently, successful architectures such as MLP-mixer~\cite{43} and Poolformer~\cite{42} have both considered the interaction between spatial and channel information in constructing context information. However, vanilla CCT focuses excessively on establishing channel information and overlooks the crucial role of spatial information in neighborhood modeling.
To address this, we develop a spatial-channel cross transformer block (SCTB) as a spatial-channel blending unit to mix full-level encoded features.
As shown in Fig.~\ref{fig3}, given the $i$-th level features ${\mathbf{I_{i}} \in \mathbb{R}^{{C_{i}} \times h \times w}}, (i= 1, 2, 3, 4)$, in which $ h= {\frac{H}{16}}, w={\frac{W}{16}}$. the procedure of SCTB can be defined as 
\begin{align}
 \mathbf{J_{\sum}} &=\text{LN}([\mathbf{I_{1}},\mathbf{I_{2}},\mathbf{I_{3}},\mathbf{I_{4}}]), \\
 \mathbf{J_{i}} &=\text{LN}(\mathbf{I_{i}}), \\
 \mathbf{P_{i}} &=\text{SSCA}({\mathbf{J_{1}}, \mathbf{J_{2}}, \mathbf{J_{3}}, \mathbf{J_{4}}, \mathbf{J_{\sum}}})+\mathbf{I_{i}}, \\
 \mathbf{O_{i}} &={\text{CFN}_{i}}(\mathbf{P_{i}}), 
\end{align}
where LN denotes the layer normalization, ${\mathbf{J_{i}} \in \mathbb{R}^{{C_{i}} \times h \times w}}, (i= 1, 2, 3, 4)$ and the concatenated tokens ${\mathbf{J_{\sum}} \in \mathbb{R}^{{C_{\sum}} \times h \times w}}$ are the five inputs of SSCA, $\mathbf{P_{i}}$ represent the outputs of SSCA, and $\mathbf{O_{i}}$ stands for the outputs of SCTB. The SSCA: Spatial-embedded single-head channel-cross attention; and CFN: Complementary feed-forward network, are separately described below.

\begin{table*}[t]
\renewcommand{\arraystretch}{1.3}
\caption{Comparisons with SOTA methods on NUAA-SIRST, NUDT-SIRST and IRSTD-1K in $IoU(\%)$, $nIoU(\%)$, $F$-$measure(\%)$, $Pd(\%)$, $Fa({10}^{-6})$.}
\label{tab1}
\setlength\tabcolsep{1.4mm}
\begin{tabular}{c|ccccc|ccccc|ccccc} \hline
\multirow{3}{*}{Method} & \multicolumn{5}{c|}{NUAA-SIRST~\cite{7}}                                           & \multicolumn{5}{c|}{NUDT-SIRST~\cite{8}}                                           & \multicolumn{5}{c}{IRSTD-1K~\cite{13}}                                           \\ \cline{2-16} 

                        & mIoU      & nIoU     & {F-measure}       & Pd              & Fa             & mIoU      & nIoU     & {F-measure}       & Pd              & Fa             & mIoU      & nIoU     & {F-measure}       & Pd              & Fa             \\\hline
Top-Hat~\cite{1}                 & 7.143     & 18.27      & {14.63}    & 79.84          & 1012          & 20.72     &  28.98    & {33.52}    & 78.41           & 166.7          & 10.06     & 7.438    &{16.02}    & 75.11           & 1432         \\
Max-Median~\cite{44}             & 4.172     & 12.31      & {10.67}    & 69.20           & 55.33         &4.197     &  3.674    & {7.635}    & 58.41           & 36.89          & 6.998     & 3.051    &{8.152}    & 65.21           & 59.73          \\
WSLCM~\cite{45}              & 1.158     & 6.835      & {4.812}    & 77.95           & 5446         & 2.283     &  3.865   & {5.987}    & 56.82           & 1309           & 3.452     & 0.678    &{2.125}    & 72.44           & 6619          \\
TTLCM~\cite{46}                 & 1.029     & 4.099     & {4.995}    & 79.09           & 5899          & 2.176     &  4.315     & {7.225}    & 62.01          & 1608         & 3.311     & 0.784    &{2.186}    & 77.39           & 6738          \\
IPI~\cite{5}                 & 25.67     &  50.17    & {43.65}    & 84.63           & 16.67         & 17.76     &  15.42    & {26.94}    & 74.49          & 41.23          & 27.92     & 20.46    &{35.68}    & 81.37           & 16.18          \\
PSTNN~\cite{70}  & 30.30   & 33.67   & 39.16    & 72.80  & 48.99
      & 14.85  & 23.57    & 35.63   & 66.13  & 44.17
      &  24.57 & 17.93   & 37.18    & 71.99  & 35.26\\

MSLSTIPT~\cite{47}           & 10.30      &  15.93    & {18.83}     & 82.13            & 1131        & 8.342     &  10.06   & {18.26}    & 47.40         & 888.1            & 11.43     & 5.932    &{12.23}    & 79.03          & 1524          \\
ACM~\cite{7}                  & 68.93     & 69.18    & {80.87}    & 91.63           & 15.23          & 61.12     & 64.40    & {75.87}    & 93.12           & 55.22          & 59.23     & 57.03    &{74.38}    & 93.27           & 65.28          \\
ALCNet~\cite{29}                  & 70.83     & 71.05    & 82.92    & 94.30           & 36.15          & 64.74     & 67.20    & {78.59}    & 94.18           & 34.61          & 60.60     & 57.14    & {75.47}    & 92.98           & 58.80          \\

RDIAN~\cite{12}                  & 68.72     & 75.39    & 81.46   & 93.54           & 43.29          & 76.28     & 79.14    & {86.54}    & 95.77           & 34.56          & 56.45     & 59.72    & {72.14}    & 88.55           & 26.63          \\
ISTDU~\cite{9}              & 75.52     & 79.73    & {86.06}    & \underline{96.58}           & 14.54          & 89.55     & 90.48    & {94.49}    & 97.67           & 13.44          & \underline{66.36}     & 63.86    & 79.58   & \underline{93.60}           & 53.10          \\
MTU-Net~\cite{11}                & 74.78     & 78.27    & {85.37}    & 93.54           & 22.36          & 74.85     & 77.54    & {84.47}      & 93.97           & 46.95          & 66.11     & 63.24    & {79.26}      & 93.27           & 36.80          \\
IAANet~\cite{69}   & 74.22    & 75.58    & 85.02    & 93.53    & 22.70 
                    & 90.22     &  92.04   & 94.88    & 97.26    & 8.32   
                   & 66.25     & 65.77    & 78.34    & 93.15    & 14.20  \\

AGPCNet~\cite{30}   & 75.69     & 76.60    & 85.26    & 96.48   & 14.99 
                          & 88.87     & 90.64    & 93.88    & 97.20    & 10.02   
                          & 66.29    & 65.23    &  79.58   & 92.83      & 13.12          \\
DNA-Net~\cite{8}               & 75.80     & 79.20    & 86.24    & 95.82           & \textbf{8.78}          & 88.19     & 88.58    & {93.73}    & \textbf{98.83}           & 9.00           & 65.90     & 66.38    & {79.44}    & 90.91           & 12.24          \\
UIU-Net~\cite{10}              & \underline{76.91}     & \underline{79.99}    & {\underline{86.95}}    & 95.82           & 14.13          & \underline{93.48}     & \underline{93.89}    & {\underline{96.63}}    & 98.31           & \underline{7.79}           & 66.15     & \underline{66.66}    &  \underline{79.63}    & \textbf{93.98}           & 22.07          \\
SCTransNet               & \textbf{77.50}     & \textbf{81.08}    & {\textbf{87.32}}    & \textbf{96.95}           & \underline{13.92}           & \textbf{94.09}     & \textbf{94.38}    & {\textbf{96.95}}    & \underline{98.62}           & \textbf{4.29}           & \textbf{68.03}     & \textbf{68.15}    & {\textbf{80.96}}    & 93.27           & \textbf{10.74} \\ \hline       
\end{tabular}
\end{table*}

\subsubsection{Spatial-embedded single-head channel-cross attention}
\label{labSSCA}
In Fig.~\ref{fig3}(a), given the five input tokens $\mathbf{J_{i}}$ and $\mathbf{J_{\sum}}$ for which LN is performed, the launching point of SSCA is to calculate the local-spatial channel similarity between single-level features and full-level concatenation features to establish global semantics.
Therefore, our SSCA employs the four input tokens $\mathbf{J_{i}}$ as queries, one concatenated token $\mathbf{J_{\sum}}$ as key and value. 
This is accomplished by utilizing $1\times 1$ convolutions to consolidate pixel-wise cross-channel context and then applying $3\times 3$ depth-wise convolutions to capture local spatial context. Mathematically, 
% It is achieved by applying $1\times 1$ convolutions to aggregate pixel-wise cross-channel context followed by $3\times 3$ depth-wise convolutions to encode local spatial context. Mathematically, 
\begin{equation}
% \label{eq1}
\begin{split}
\mathbf{Q_{i}}={W^{Q}_{di}}{W^{Q}_{pi}}\mathbf{J_{i}},~\mathbf{K}={W^{K}_{d}}{W^{K}_{p}}\mathbf{J_{\sum}},~\mathbf{V}={W^{V}_{d}}{W^{V}_{p}}\mathbf{J_{\sum}},
\end{split}
\end{equation}
where ${W^{(\cdot)}_{pi}} \in \mathbb{R}^{{C_{i}} \times 1 \times 1}$ and ${W^{(\cdot)}_{p}} \in \mathbb{R}^{{C_{\sum}} \times 1 \times 1}$ are the $1\times 1$ point-wise convolution, ${W^{(\cdot)}_{di}}\in \mathbb{R}^{{C_{i}} \times 3 \times 3}$ and ${W^{(\cdot)}_{d}}\in \mathbb{R}^{{C_{\sum}} \times 3 \times 3}$ are the 3×3 depth-wise convolution. 
Next, we reshape ${\mathbf{Q_{i}}}\in \mathbb{R}^{{C_{i}} \times h \times w}$, ${\mathbf{K}}\in \mathbb{R}^{{C_{\sum}} \times h \times w}$, and ${\mathbf{V}}\in \mathbb{R}^{{C_{\sum}} \times h \times w}$ to $\mathbb{R}^{{C_{i}} \times hw}$, $\mathbb{R}^{{C_{\sum}} \times hw}$ and $\mathbb{R}^{{C_{\sum}} \times hw}$, separately. Our SSCA process is defined as
\begin{align}
\label{eq2}
& {\mathbf{CA_{i}}}={W_{pi}}~\text{CrossAtt}(\mathbf{Q_{i}}, \mathbf{K}, \mathbf{V}), \\
& \text{CrossAtt}(\mathbf{Q_{i}}, \mathbf{K}, \mathbf{V})={\mathbf{{A}_{i}}}{\mathbf{V}}=\text{Softmax}\left\{\mathcal{I}(\frac{{\mathbf{Q_{i}}}~{\mathbf{K^{T}}}}{\lambda})\right\}{\mathbf{V}},
\end{align}
where ${\mathbf{CA_{i}}} \in \mathbb{R}^{{C_{i}} \times h \times w}$ are the output of SSCA,
${\mathbf{{A}_{i}}} \in \mathbb{R}^{{C_{i}} \times {C_{\sum}}}$ represent different level covariance-based attention maps, $\mathcal{I}$ denotes the instance normalization operation~\cite{66}, and~$\lambda$ is an optional temperature factor defined by $\lambda = \sqrt{C_{\sum}}$.
% and~$\lambda$ is an optional temperature factor defined by $\lambda = \sqrt{C_{\sum}}$.
Notably, we differ from the common channel-cross attention under two further aspects: Our patches are without positional encoding, and we use a single head to learn the attention matrix.
These strategies will be compared for their efficacy in detail in the ablation study~\ref{Abla SSCA}.

\begin{figure}[t!]
   \centering
    \includegraphics[width=8cm]{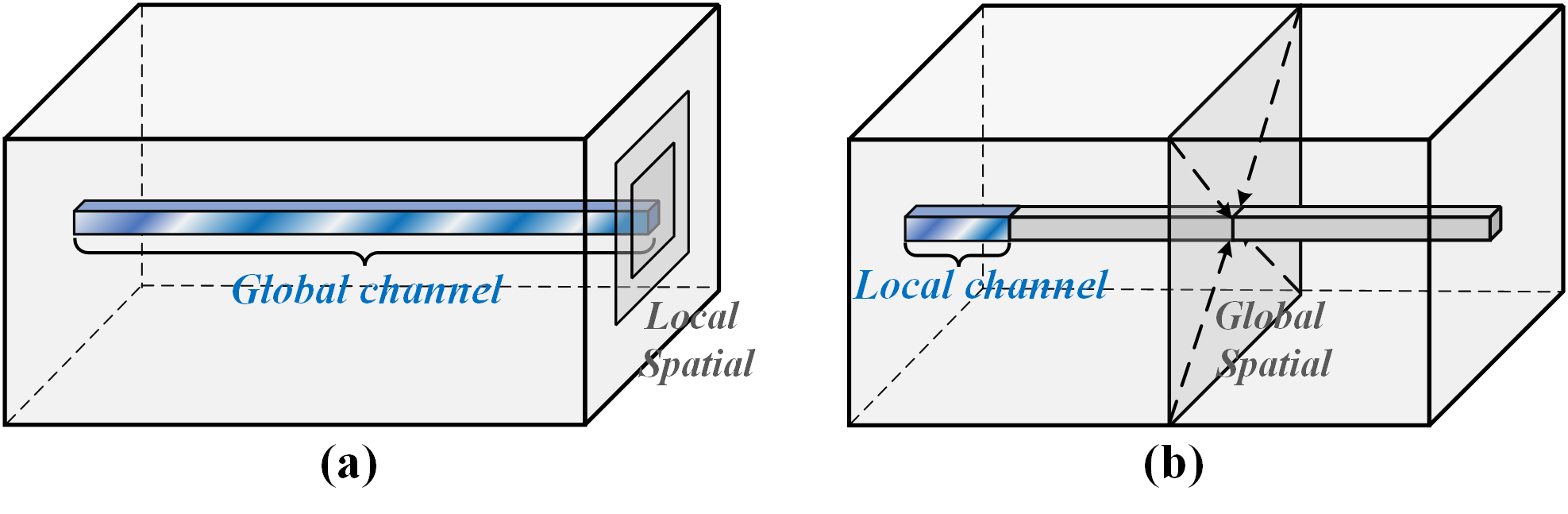}
    \caption{Information enhancement from different perspectives: (a) the local spatial and global channel (LSGC) paradigms; (b) the global spatial and local channel (GSLC) paradigms. Our CFN integrates both of these information enhancement methods internally.}
    \label{fig4}
\end{figure}

\subsubsection{Complementary Feed-forward Network}
\label{labCFN}
As shown in Fig.~\ref{fig4}(a), 
% previous studies~\cite{14},~\cite{48},~\cite{22} usually introduce single-scale depth-wise convolutions into the regular feed-forward network to improve locality.
previous studies~\cite{14},~\cite{48},~\cite{22} always incorporate single-scale depth-wise convolutions into the standard feed-forward network to enhance local focus.
More recently, state-of-the-art MSFN~\cite{67} incorporates two paths with depth-wise convolution using different kernel sizes to enhance the multi-scale representation.
However, the above approaches are limited to a local spatial global channel paradigm of feature representation.
In fact, global spatial and local channel information~(Fig.~\ref{fig4}(b)) is equally important~\cite{63}. Hence, we design a CFN, which combines the advantages of both feature representations.

In Fig.~\ref{fig3}(b), given an input tensor ${\mathbf{X_{i}}} \in \mathbb{R}^{{C_{i}}\times h \times w }$, CFN first models multi-scale LSGC information. Specifically, after the layer normalization, 
\replaced{CFN utilizes $1\times 1$ convolution to increase the channel dimension in the ratio of $\eta$ and splits the feature map equally into two branches. Subsequently, $3 \times 3$ and $5\times 5$ depth-wise convolutions are employed to enhance the local spatial information.}
{CFN utilizes $1\times 1$ convolution to increase the channel dimension by a factor of $\eta$, and equally divides the feature map into two branches and enhances the local spatial information using $3 \times 3$ and $5\times 5$ depth-wise convolution, respectively.}
This is followed by channel concatenating the multi-scale features and restoring them to their original dimensions. The above process can be defined as
\begin{align}
\label{eq3}
& {{\mathbf{X_{3\times3}}}, {\mathbf{X_{5\times5}}}}= \text{Chunk}({f^{c}_{1\times1}}(LN({\mathbf{X_{i}}}))), \\
& {\mathbf{X_{sc}}}={f^{c}_{1\times1}}[\delta({f^{dwc}_{3\times3}({\mathbf{X_{3\times3}}})}), \delta({f^{dwc}_{5\times5}}({\mathbf{X_{5\times5}}}))], 
\end{align}
where ${f^{c}_{1\times1}}$ denotes $1\times1$ convolution, $f^{dwc}_{3\times3}$ and $f^{dwc}_{5\times5}$ represent $3\times3$ and $5\times5$ depth-wise convolutions. Here, Chunk($\cdot$) denotes dividing the feature vector into two equal parts along the channel dimension.

\begin{table}[t]
\caption{Comprehensive evaluation metrics with competitive algorithms.}
\label{tab2}
\centering
\renewcommand{\arraystretch}{1.3}
\setlength\tabcolsep{1.6mm}
\begin{tabular}{cccccc}
\hline \hline
Model   & Params (M) & Flops (G) & IoU &nIoU &F-measure\\ \hline
DNA-Net~\cite{8}  & \textbf{4.697}        &  \textbf{14.26}           & 80.23 &82.59 &88.60      \\
UIU-Net~\cite{10}  & 50.54                 & 54.42           & \underline{82.40} & \underline{86.12} & \underline{90.35}     \\ 
SCTransNet        & \underline{11.19 }                & \underline{20.24}   & \textbf{83.43} & \textbf{86.86} & \textbf{90.96}\\ \hline
\end{tabular}
\end{table}

Next, CFN constructs the GSLC information. Because of the varying resolution of the small target detection image inputs in the test stage, we first use the global average pooling (GAP) of spatial dimensions to approximate the total spatial information of the features instead of using computationally intensive spatial MLPs to precisely compute the global spatial information~\cite{62}. We then employ a one-dimensional convolution with a kernel size of $3$ to capture the local channel information of the spatially compressed feature as follows
\begin{align}
\label{eq4}
{\mathbf{X_{o}}}={f^{1D}_{3}}({\text{GAP}_{2D}}({\mathbf{X_{sc}}}))\odot{\mathbf{X_{sc}}}+{\mathbf{X_{i}}}, 
\end{align}
where $\odot$ is the broadcasted Hadamard product. 
By incorporating complementary spatial and channel information, CFN enriches the representation of features in terms of the target's localization and the background's global continuity.

\section{Experiments and Analysis}
\subsection{Evaluation metrics} 
We compare the proposed SCTransNet with the state-of-the-art (SOTA) methods using several standard metrics.

1) \textit{Intersection over Union (IoU)}: IoU is a pixel-level evaluation metric defined as
\begin{equation}
IoU =\frac{A_{i}}{A_{u}}=\frac{\sum_{i=1}^{N}{TP[i]}}{\sum_{i=1}^{N}(T[i]+P[i]-TP[i])},
\end{equation}
where ${{A}_{i}}$ and ${{A}_{u}}$ denote the size of the intersection region and union region, respectively. \textit{N} is the number of samples, \textit{TP}[$\cdot$] denotes the number of true positive pixels, \textit{T}[$\cdot$] and \textit{P}[$\cdot$] represent the number of ground truth and predicted positive pixels, respectively.

2) \textit{Normalized Intersection over Union (nIoU)}: nIoU is the normalized version of IoU~\cite{7}, given as
\begin{equation}
%\label{deqn_ex1}
nIoU =\frac{1}{N}\sum_{i=1}^{N}\frac{TP[i]}{T[i]+P[i]-TP[i]}.
\end{equation}

3) \textit{F-measure (F)}: It evaluates the miss detection and  false alarms at  pixel-level, given as
\begin{equation}
F =\frac{2\times{Prec}\times{Rec}}{{Prec}+{Rec}},
\end{equation}
where ${Prec}$ and ${Rec}$ denote the precision rate and recall rate respectively.

\begin{figure*}[t]
    \centering
    \includegraphics[width=17.9cm]{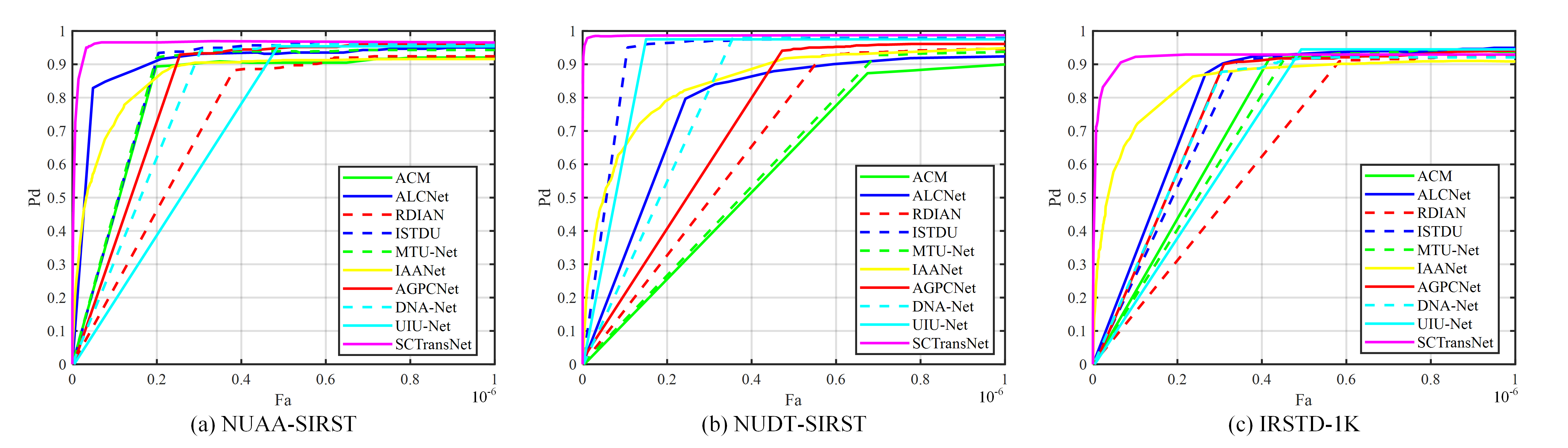}
    \caption{ROC curves of different methods on the NUAA-SIRST, NUDT-SIRST, and IRSTD-1K dataset. Our SCTransNet can achieve the highest ${{P}_{d}}$ at very low ${{F}_{a}}$.
    }
    \label{fig5}
\end{figure*}

\begin{table*}[t]
\centering
\caption{The Area Under Curve (AUC) with different thresholds of the SOTA methods on the NUAA-SIRST, NUDT-SIRST, and IRSTD-1K datasets.}
\label{tab3}
\renewcommand{\arraystretch}{1.3}
\setlength\tabcolsep{1.6mm}
\begin{tabular}{cccccccccccc}  \hline  \hline
Dataset                   & Index   & ACM    & ALCNet   & RDIAN  & ISTDU  & MTU-Net & IAANet & AGPCNet & DNA-Net & UIU-Net & SCTransNet \\  \hline
\multirow{2}{*}{NUAA-SIRST~\cite{7}}     & ${\mathrm{AUC}_{{F}_{a} = 0.5}}$  & 0.7223 & \underline{0.8618} & 0.5461 & 0.7515 & 0.7457 &0.8081   &0.6953 & 0.6582 & 0.4854 & \textbf{0.9539}    \\
                                & ${\mathrm{AUC}_{{F}_{a} = 1}}$    & 0.8180 & \underline{0.9025} & 0.7321 & 0.8579 & 0.8437 &0.8614   & 0.8262 & 0.8098 & 0.7197 & \textbf{0.9589}     \\  \hline
\multirow{2}{*}{NUDT-SIRST~\cite{8}}     & ${\mathrm{AUC}_{{F}_{a} = 0.5}}$  & 0.4392 & 0.6321 & 0.4630 & \underline{0.8635} & 0.4640 &0.7569   &0.5038 & 0.6300 & 0.8275 & \textbf{0.9853}     \\
                                & ${\mathrm{AUC}_{{F}_{a} = 1}}$    & 0.5865 & 0.7716 & 0.6695 &\underline{0.9211}  & 0.6064 & 0.8463   & 0.7306 & 0.8072 & 0.9013 & \textbf{0.9863}     \\  \hline
\multirow{2}{*}{IRSTD-1K~\cite{13}} & ${\mathrm{AUC}_{{F}_{a} = 0.5}}$  & 0.5374 & 0.6606 & 0.4545 & 0.6014 & 0.5018 & \underline{0.7862}   & 0.6211  & 0.6162 & 0.4749 & \textbf{0.9107}     \\
                                & ${\mathrm{AUC}_{{F}_{a} = 1}}$    & 0.7366 & 0.8006 & 0.6480 & 0.7687 & 0.7198 & \underline{0.8456}   & 0.7752 & 0.7684 & 0.7099 & \textbf{0.9200}     \\  \hline
\end{tabular}
\end{table*}

4) \textit{Probability of Detection (${{P}_{d}}$)}: \textit{${{P}_{d}}$} is the ratio of correctly predicted targets \textit{N$_{\mbox{\scriptsize pred}}$} and all targets \textit{N$_{\mbox{\scriptsize all}}$}, given as
\begin{equation}
P_{d} = \frac{N_{pred}}{N_{all}}.
\end{equation}
Following~\cite{8}, if the deviation of target centroid is less than 3, we consider the target correctly predicted.

5) \textit{False-Alarm Rate (${{F}_{a}}$)}: \textit{${{F}_{a}}$} is the ratio of false predicted target pixels ${N}_{false}$ and all the pixels in the image ${P}_{all}$, given as
\begin{equation}
%\label{deqn_ex1}
F_{a}=\frac{N_{false}}{P_{all}}. 
\end{equation}

In addition to the fixed-threshold evaluation methods, we also utilize Receiver Operation Characteristics (ROC) curves to comprehensively evaluate the models. ROC is used to describe the changing trends of  \textit{${{P}_{d}}$} under varying \textit{${{F}_{a}}$}.

\begin{figure*}[t]
    \centering
    \includegraphics[width=17.4cm]{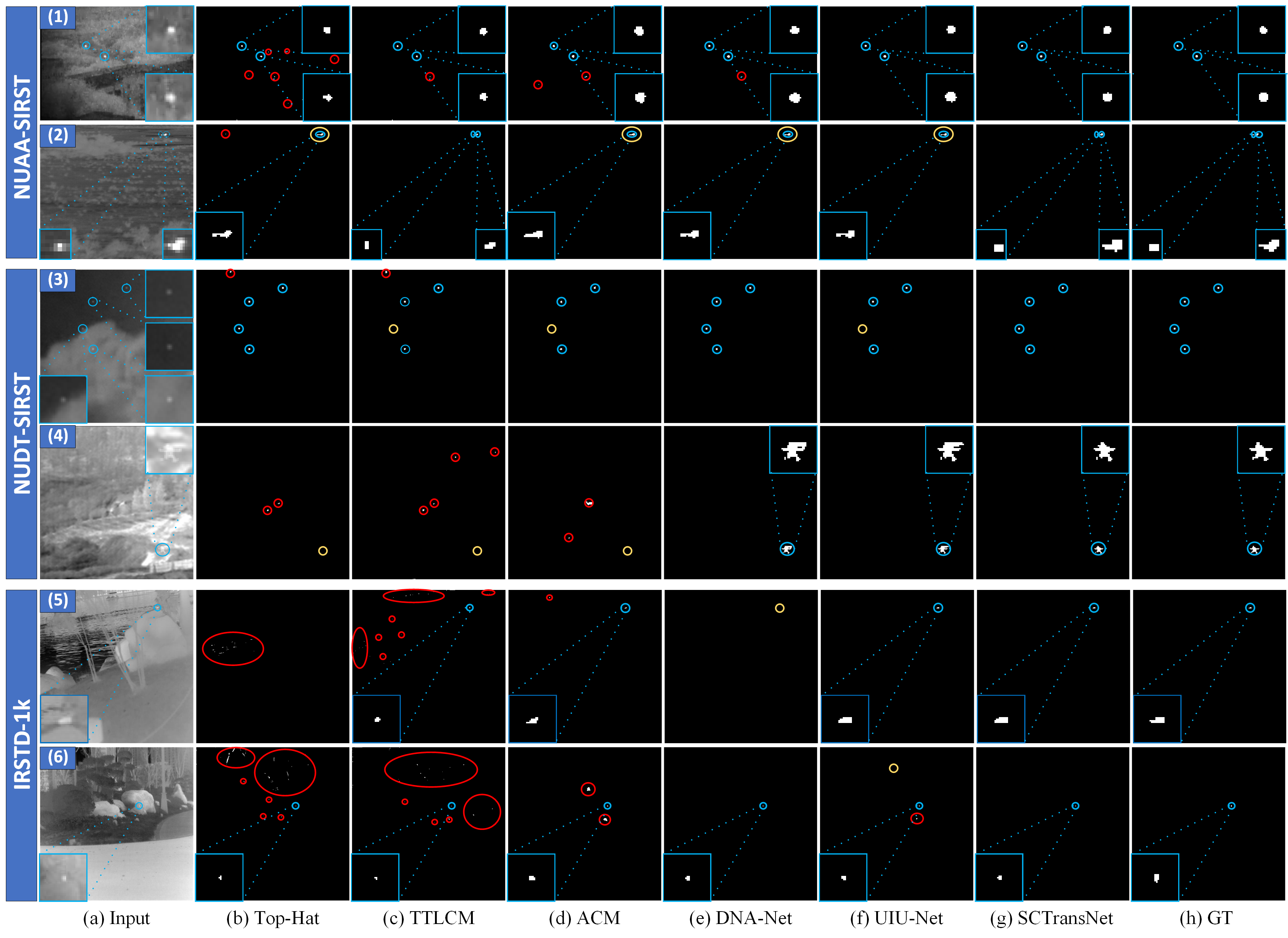}
    \caption{Visual results obtained by different IRSTD methods on the NUAA-SIRST, NUDT-SIRST, and IRSTD-1K datasets. Circles in blue, yellow, and red represent correctly detected targets, miss detection, and false alarms, respectively.}
    \label{fig6}
\end{figure*}

\subsection{Experiment settings}
\noindent {\textbf{Datasets:}} 
In our experiments, we utilized three public datasets, namely; NUAA-SIRST~\cite{7}, NUDT-SIRST~\cite{8}, and IRSTD-1K~\cite{13}, which consist of 427, 1327, and 1000 images, respectively. 
% We follow~\cite{8} to split the training and test sets of NUAA-SIRST and NUDT-SIRST, 
We adopt the method used by ~\cite{8} to partition the training and test sets of NUAA-SIRST and NUDT-SIRST,
and \cite{13} for splitting the IRSTD-1K. Hence, all splits are standard. 

\noindent {\textbf{Implementation Details:}}
We employ U-Net with four RBs as our detection backbone~\cite{11}, the number of downsampling layers is 4, and the basic width is set to 32. The kernel size and stride size $P$ for patch embedding is 16, the number of SCTB is 4, and the channel expansion factor $\eta$ in CFN is 2.66.
Our SCTransNet does not use any pre-trained weights for training, 
every image undergoes normalization and random cropping into 256×256 patches.
To avoid over-fitting, we augment the training data through random flipping and rotation.
We initialized the weights and bias of our model using the Kaiming initialization method~\cite{68}.
The model is trained using the BCE loss function and optimized by the Adam optimizer with the initial learning rate of 0.001, and the learning rate is gradually decreased to $1\times{{10}^{-5}}$ using the Cosine Annealing strategy.
The batch size and epoch are set as 16 and 1000, respectively.
Following~\cite{7},~\cite{29},~\cite{8}, the fixed threshold to segment the salient map is set to 0.5. The proposed SCTransNet is implemented with PyTorch on a single Nvidia GeForce 3090 GPU, an Intel Core i7-12700KF CPU, and 32 GB of memory. The training process took approximately 24 hours.

\noindent {\textbf{Baselines:}}
To evaluate the performance of our method, 
\replaced{we compare SCTransNet to the SOTA IRSTD methods, specifically, seven well-established traditional methods}
{we compare SCTransNet to the SOTA IRSTD methods. Specifically, we compare it with seven well-established traditional methods} (Top-Hat\cite{1}, Max-Median~\cite{44}, WSLCM~\cite{45}, TLLCM~\cite{46}, IPI~\cite{5}, MSLSTIPT~\cite{47}), and nine learning-based methods (ACM~\cite{7}, ALCNet~\cite{29}, RDIAN~\cite{12}, ISTDU~\cite{9}, IAANet~\cite{69}, AGPCNet~\cite{30}, DNA-Net~\cite{8},  UIU-Net~\cite{10}, and MTU-Net~\cite{11}) on the NUAA-SIRST, NUDT-SIRST and IRSTD-1K datasets. To guarantee an equitable comparison, we retrained all the learning-based methods using the same training datasets as our SCTransNet, and following the original papers, adopted their fixed thresholds. Open-source implementations of most techniques can be found at \url{https://github.com/XinyiYing/BasicIRSTD} and \url{https://github.com/xdFai/SCTransNet}.

\begin{figure*}[t]
    \centering
    \includegraphics[width=17.8cm]{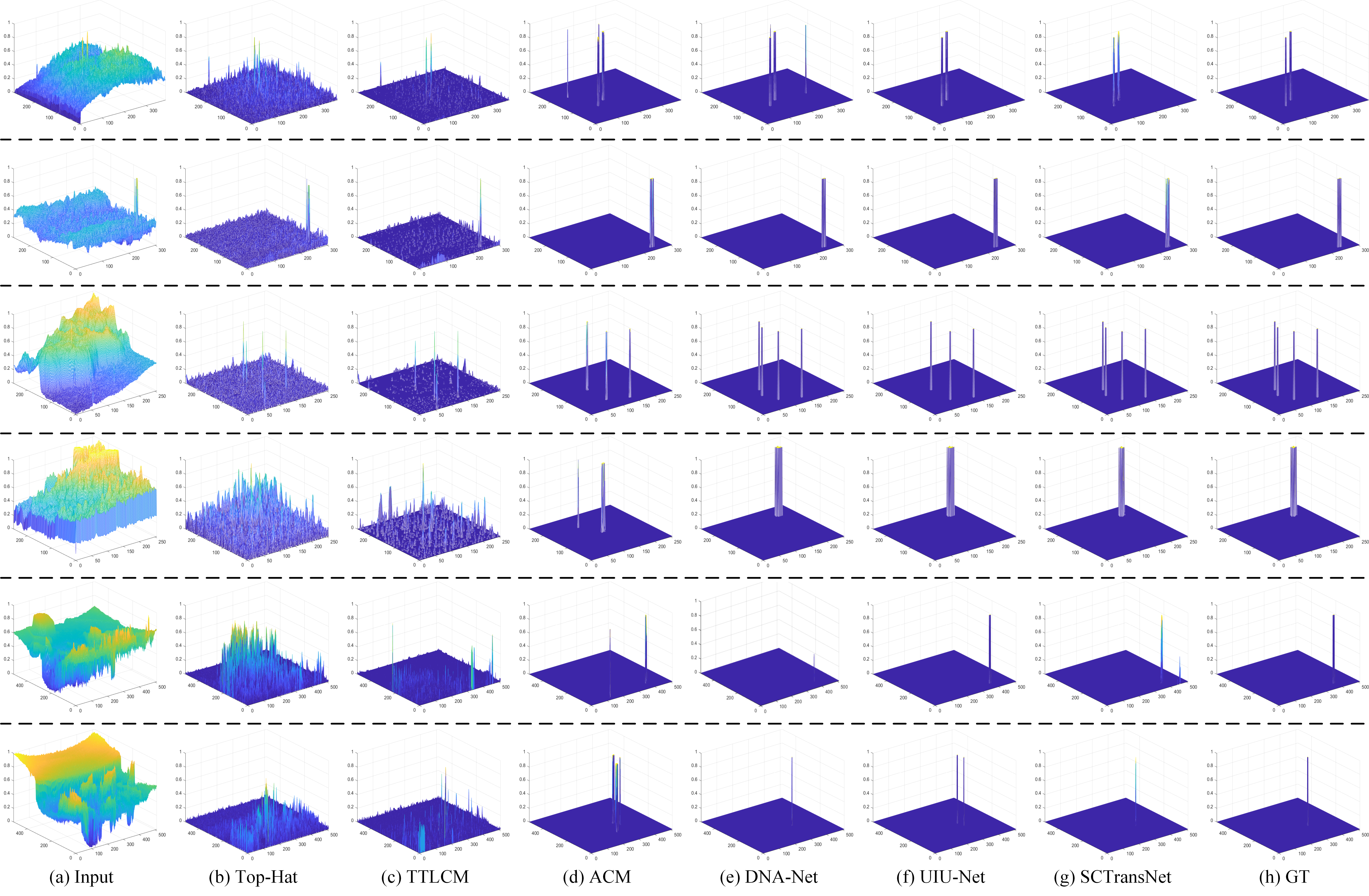}
    %%\vspace{-3mm}
    \caption{3D visualization of salient maps of different methods on 6 test images.}
    %%\vspace{-1mm}
    \label{fig7}
\end{figure*}

\subsection{Quantitative Results}
Quantitative results are shown in Table~\ref{tab1}. In general, the learning-based methods significantly outperform the conventional algorithms in terms of both target detection accuracy and contour prediction of targets.
Meanwhile, our method outperforms all other algorithms. In the three metrics of \textit{IoU}, \textit{nIoU} and \textit{F}-\textit{measure}, SCTransNet stands considerably ahead on all three public datasets. This indicates that our algorithm possesses a strong ability to retain target contours and can discern pixel-level information differences between the target and the background.
We also note that even though SCTransNet does not obtain optimal ${{P}_{d}}$ and ${{F}_{a}}$, e.g., DNA-Net's ${{P}_{d}}$ is higher than ours by only 0.2 in the NUDT-SIRST, whereas our target detection false alarms are over twice as low as DNA-Net's. 
This demonstrates that our algorithm achieves a superior balance between false alarms and detection accuracy, as indicated by the remarkably high composite metric, \textit{F}-\textit{measure}.
Next, we comprehensively compare the present algorithm with the most competitive deep learning methods, DNA-Net and UIU-Net.
Table~\ref{tab2} gives the average metrics of the different algorithms on the three data, and \textcolor{black}{we can observe that SCTransNet has acceptable parameters at the highest performance} 
and outperforms the powerful UIU-Net.

Fig.~\ref{fig5} displays the ROC curves of various competitive learning-based algorithms. It is evident that the ROC curve of SCTransNet outperforms all other algorithms.
For instance, by appropriately selecting a segmentation threshold, SCTransNet achieves the highest detection accuracy while maintaining the lowest false alarms in the NUAA-SIRST and NUDT-SIRST datasets.

Table~\ref{tab3} presents the Area Under Curve (AUC) of Fig.~\ref{fig5} in two different thresholds: ${{F}_{a}=0.5\times{10}^{-6}}$ and ${{F}_{a}=1\times{10}^{-6}}$.
It can be seen that our method consistently achieves optimal detection performance across various false alarm rates.
Meanwhile, while undergoing the same continuous threshold change, the curve of our method is more continuous and rounded compared to other methods. This observation suggests that SCTransNet showcases exceptional tunable adaptability.

\subsection{Visual Results}
The qualitative results of the seven representative algorithms in the NUAA-SIRST, NUDT-SIRST, and IRSTD-1K datasets are given in Fig.~\ref{fig6} and Fig.~\ref{fig7}.
Among them, conventional algorithms such as Top-Hat and TTLCM frequently yield a high number of false alarms and missed detections. Furthermore, even in cases where the target is detected, its contour is often unclear, hindering further accurate identification of the target type.
In the learning-based algorithms, our method achieves precise target detection and effective contour segmentation. As illustrated in Fig.~\ref{fig6}(2), our method successfully distinguishes between two closely located targets, whereas other deep learning methods tend to merge them into a single target.
This suggests that our method discriminates each element in the image accurately.
In Fig.~\ref{fig6}(4), only our method accurately separates the shape of the unmanned aerial vehicle (UAV) from the mountain range. This is because our method not only learns the target's features but also constructs high-level semantic information about the backgrounds, thereby accurately capturing the overall continuity of the background. 
In Fig.~\ref{fig6}(6), except for the present method and DNA-Net, the remaining methods produce false alarms on the stone in the grass. This can be attributed to their limitation in only constructing local contrast information and lack of establishing long-distance dependence on the image.

\begin{table}[t]
\centering
\caption{Based on U-Net, ablation study of the \replaced{residual blocks}{ResBlocks} (RBs), deep supervision (DS), SSCA, CFN, and CCA module in average $IoU(\%)$, $nIoU(\%)$, $F$-$measure(\%)$ on NUAA-SIRST, NUDT-SIRST, and IRSTD-1K.}
\label{tab4}
\renewcommand{\arraystretch}{1.3}
\setlength\tabcolsep{1.1mm}
\begin{tabular}{ccccccccc}\hline\hline
 U-Net & +RBs &+DS &+SSCA & +CFN & +CCA & IoU & nIoU & F-measure   \\\hline
\ding{51}  & \ding{55}  & \ding{55}    & \ding{55}     & \ding{55}  & \ding{55}    & 75.29 &78.60 &86.36 \\\hline
\ding{51}  &\ding{51}   & \ding{55}    & \ding{55}     & \ding{55}  & \ding{55}    & 77.07 &80.13 &87.05 \\\hline
\ding{51}  & \ding{51}     & \ding{51}     & \ding{55}     & \ding{55}  & \ding{55}    & 77.73 &80.78 &87.47 \\\hline
\ding{51}  & \ding{51}    & \ding{51}     & \ding{51}      & \ding{55}  & \ding{55}    & 82.39 &85.71 &90.34 \\\hline
\ding{51}  & \ding{51}    & \ding{51}     & \ding{51}      & \ding{51}   & \ding{55}    & 82.89 &86.28 &90.66 \\\hline
\ding{51}  & \ding{51}     & \ding{51}     & \ding{51}      & \ding{51}   & \ding{51}     & 83.43 & 86.86 &90.96 \\ \hline 
\end{tabular}
\end{table}

\begin{table}[b]
\centering
\caption{\added{Based on UCTransNet, ablation study of the RBs, DS, skip connections (SKs) and SCTB, reporting average $IoU(\%)$, $nIoU(\%)$, $F$-$measure(\%)$ on NUAA-SIRST, NUDT-SIRST, and IRSTD-1K. Note that, we replace the CCT in UCTransNet using the proposed SCTB.}}
\label{tab5}
\renewcommand{\arraystretch}{1.3}
\setlength\tabcolsep{0.8mm}
\begin{tabular}{cccccccc}\hline\hline
UCTransNet & +RBs &+DS & +SKs & SCTB r/ CCT & IoU & nIoU & F-measure   \\\hline
\ding{51}  & \ding{55}  & \ding{55}    & \ding{55}     & \ding{55}   & 78.78 &81.56 &87.80 \\\hline
\ding{51}  &\ding{51}   & \ding{55}    & \ding{55}     & \ding{55}   & 79.95 &82.97 &88.45 \\\hline
\ding{51}  & \ding{51}  & \ding{51}    & \ding{55}     & \ding{55}   & 81.47 &83.89 &88.92 \\\hline
\ding{51}  & \ding{51}  & \ding{51}    & \ding{51}     & \ding{55}   & 82.03 &84.98 &89.54 \\\hline
\ding{51}  & \ding{51}  & \ding{51}    & \ding{51}     & \ding{51}   & 83.43 &86.86 &90.66 \\\hline
\end{tabular}
\end{table}

\subsection{Ablation Study}

\added{In this section, we first employ two baselines to demonstrate the effectiveness of SCTransNet.}
\begin{itemize}
    \item \textbf{U-Net}: \added{We incrementally incorporate the residual blocks (RBs), deep supervised (DS), SSCA, CFN, and CCA into the baseline U-Net to validate the effectiveness of the above modules for infrared small target detection. The results are presented in Table~\ref{tab4}. We observe that the algorithm's performance improves consistently with the inclusion of the aforementioned modules. In particular, the SSCA module significantly enhances the $IoU$, $nIoU$, and $F$-$measure$ value of the algorithm by 4.66\%, 4.93\%, and 2.87\%, respectively. This effectively demonstrates the effectiveness of the full-level information modeling of the IR small target.}
    \item \textbf{UCTransNet}: \added{We incrementally incorporate the RBs, DS, and skip connections (SKs), and use the proposed SCTB to replace CCT in the baseline UCTransNet to validate the effectiveness of these modules. As shown in Table~\ref{tab5}, these modules consistently enhance the algorithm's performance. Particularly, the proposed SCTB improves the $IoU$, $nIoU$, and $F$-$measure$ value of the algorithm by 1.40\%, 1.88\%, and 1.12\%, respectively, compared to the primitive CCT. This demonstrates the proposed SCTB can more effectively enhance the semantic difference between IR small targets and backgrounds than CCT.}
\end{itemize}
\deleted{
In this section, we incorporate the ResBlocks (RBs), deep supervised (DS), SSCA, CFN, and CCA into the baseline U-Net to validate the effectiveness of the above modules for infrared small target detection. The results are presented in Table~\ref{tab4}. We observe that the algorithm's performance improves consistently with the inclusion of the aforementioned modules. In particular, the SSCA module significantly enhances the $IoU$, $nIoU$, and $F$-$measure$ value of the algorithm by 4.66\%, 4.93\%, and 2.87\%, respectively. This effectively demonstrates the effectiveness of the full-level information modeling of the target.
}

Next, we will delve into a detailed discussion of the proposed SCTB, SSCA and CFN, and compare the adopted CCA block with other feature fusion approaches implemented in IRSTD.

\begin{figure}[t]
    \centering
    \includegraphics[width=8.8cm]{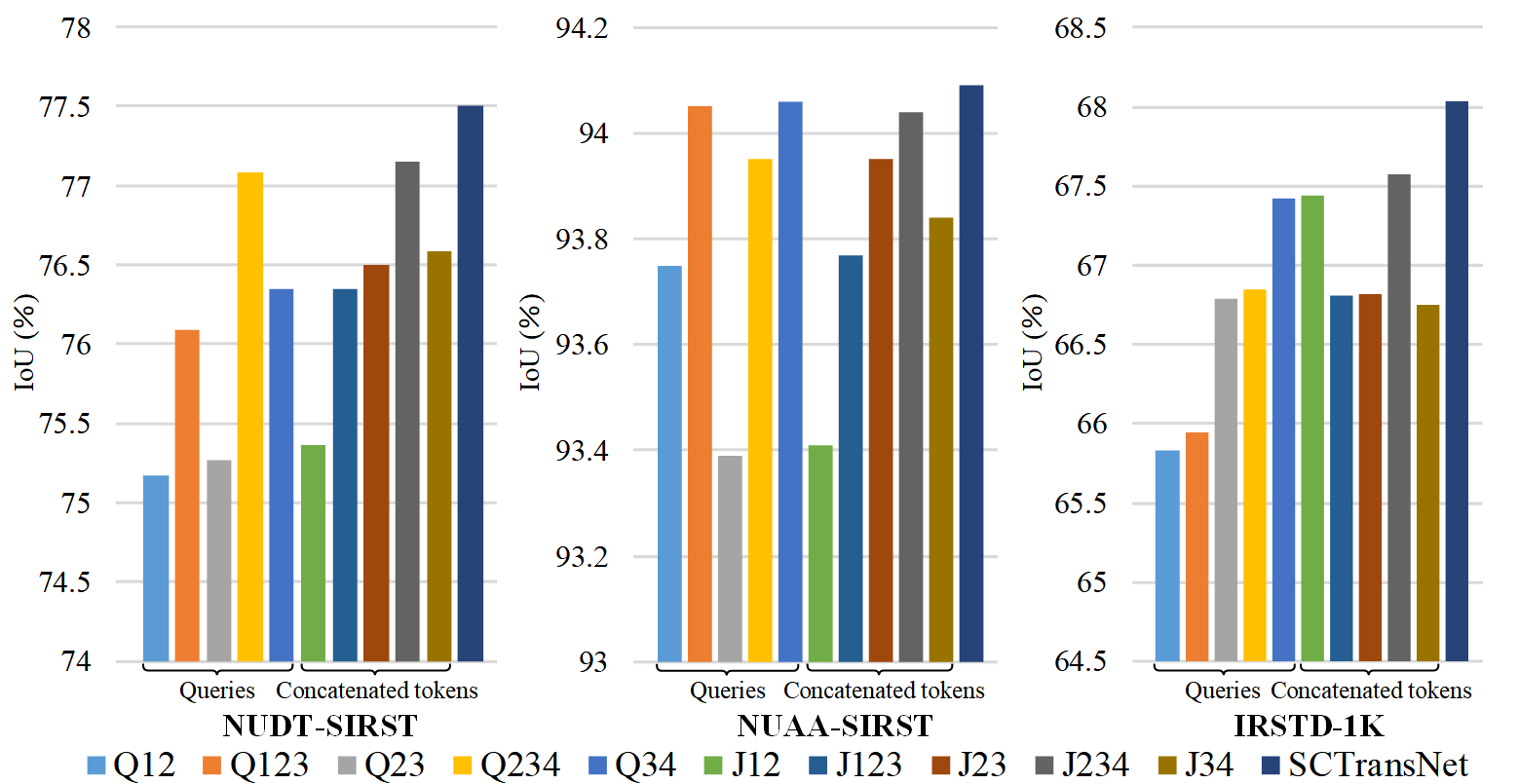}
    \caption{Ablation on level of queries (Q) and composition of concatenated feature (J) on NUAA-SIRST, SUDT-SIRST and IRSTD-1K.}
    \label{fig12}
\end{figure}

\subsubsection{The Spatial-channel Cross Transformer Block}

In the proposed SCTransNet, a primary idea is utilizing SCTB to mix and redistribute the output features of the full-stage encoders \added{to predict contextual information about the small target and backgrounds}. Since the network is encoded four times, the number of queries (Q) is set to 4, and both keys (K) and values (V) are formed by mapping the concatenated features (J) of the complete 4-level features. In this section, we will discuss different levels of Q and the composition of J to illustrate the importance of full-level feature modeling.

Fig.~\ref{fig12} presents the ablation results for the level of Q and composition of J across three datasets.
Note that when changing Q, J is composed of full-level features, and likewise, Q is the full-level feature input when varying J.
The experimental results for Q indicate significant differences in the information learned by the neural network from different levels of features.
Queries with higher and more comprehensive levels (Q123, Q234, Q34) encompass rich image semantics, thus achieving higher performance. 
The model performs best when fed with full-level Q inputs (SCTransNet), thus validating our motivation.
Similarly, the experimental results for J suggest that selecting complete channel information allows queries to capture more accurate key features, thereby improving the performance of IRSTD.

\subsubsection{The Spatial-embedded Single-head Channel-cross Attention}
\label{Abla SSCA}
To demonstrate the efficacy of the proposed SSCA, we present multi-head cross-attention~\cite{23} (MCA, a typical full-level information interaction structure \replaced{in UCTransNet}{ for medical image segmentation}) and three network structure variants: SSCA with positional encoding (\textit{SSCA w PE}), SSCA with multi-head (\textit{SSCA w MH}), and SSCA without spatial-embedding (\textit{SSCA w/o SE}), respectively.
\begin{itemize}
    \item \textbf{SSCA w PE}: We incorporate positional encoding during the patch embedding stage. To accommodate test images of different sizes, we employ interpolation to scale the position-coding matrix, ensuring the proper functioning of the algorithm.
    \item \textbf{SSCA w MH}: We use a typical multi-head cross-attention mechanism to replace the single-head cross-attention mechanism in SSCA to verify the effectiveness of the single-head strategy for extracting limited features from the IR small targets.   
    \item \textbf{SSCA w/o SE}: To validate the effectiveness of local spatial information coding, we eliminate the depth-wise convolution in the QKV matrix generation process in SCTB.
\end{itemize}

\begin{table}[t]
\centering
\caption{$IoU(\%)$/$nIoU(\%)$/$F$-$measure(\%)$ values achieved by variants of SSCA and MCA on NUAA-SIRST, NUDT-SIRST and IRSTD-1K.}
\label{tab6}
\renewcommand{\arraystretch}{1.3}
\setlength\tabcolsep{1.1mm}
\begin{tabular}{cccc}    \hline\hline
\multirow{2}{*}{Model}    & \multicolumn{3}{c}{Dataset} \\  \cline{2-4} 
                         & NUAA-SIRST            & NUDT-SIRST        & IRSTD-1K     \\ \hline
MCA~\cite{23}   & 74.72/78.35/85.53       & 93.07/93.61/96.41     & 65.60/66.57/79.22  \\ 
SSCA w PE     & 77.10/79.88/87.07       & 94.03/94.25/96.93     & 66.01/65.29/79.52  \\
SSCA w MH     &  76.35/79.56/86.59      & 93.72/94.13/96.76     & 67.08/67.55/80.30  \\
SSCA w/o SE   & 76.40/79.19/86.62       & 93.23/93.49/96.50     & 66.10/65.48/79.59  \\
SSCA            & \textbf{77.50/81.08/87.32}   & \textbf{94.09/94.38/96.95}  & \textbf{68.03/68.15/80.96}  \\
\hline
\end{tabular}
\end{table}

\begin{figure}[t]
    \centering
    \includegraphics[width=8.8cm]{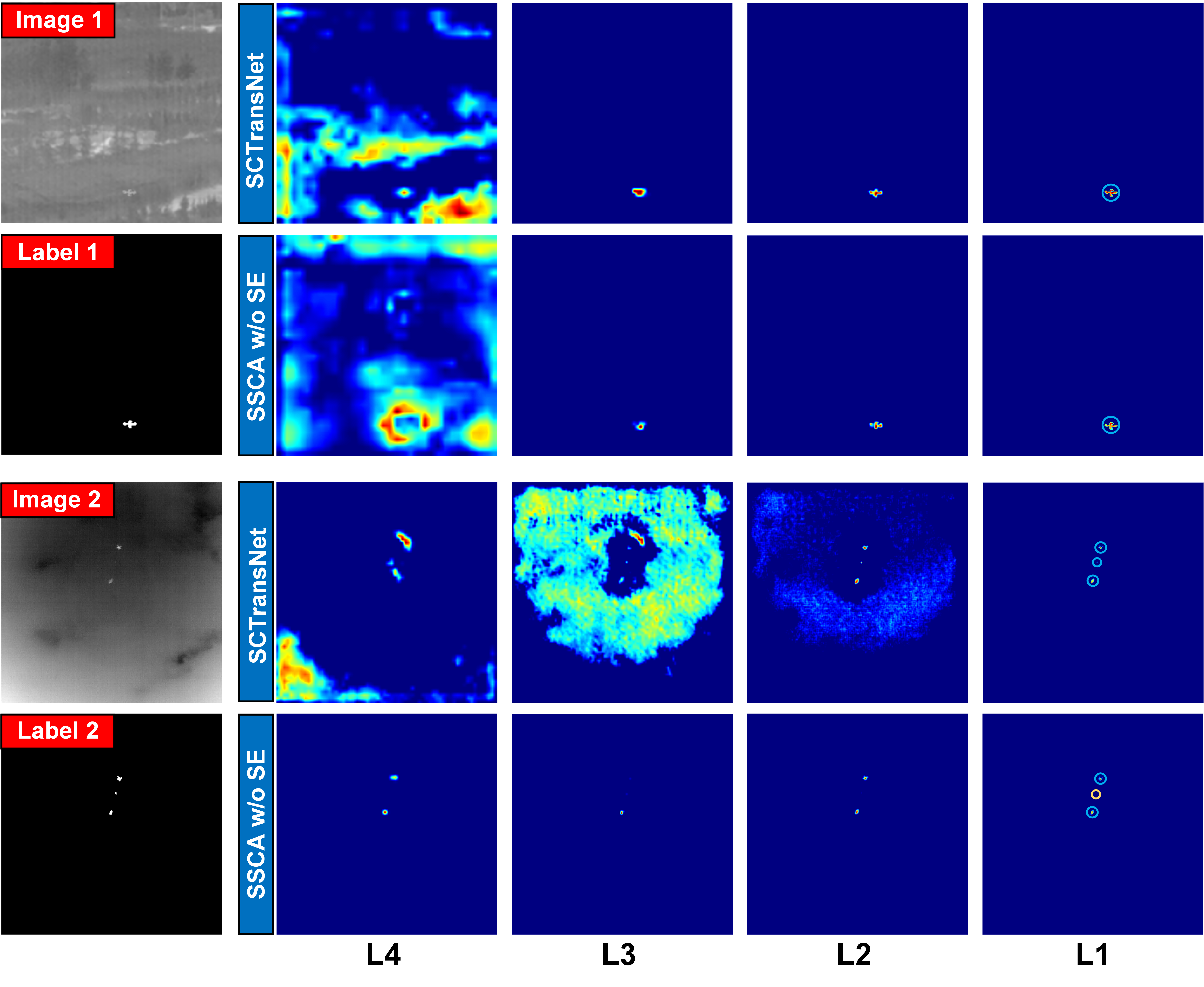}
    \caption{Visualization map of SCTransNet and SSCA w/o SE. The feature maps from the deep layers of SCTransNet have an accurate representation of the localized region of the target from the background, and accurate segmentation results are obtained at the output layer.}
    \label{fig8}
\end{figure}

As illustrated in Table~\ref{tab6}, our SSCA has higher $IoU$, $IoU$, and $F$-$measure$ values than the MCA  and the variant \textit{SSCA w PE} on three datasets.
This suggests that SCTransNet can better perceive the information difference between small targets and complex backgrounds than MCA through comprehensive information interaction.
It also illustrates that absolute positional encoding is not suitable for IRSTD tasks.
This is due to the scaling of the position-embedding matrix in variable-size image inputs, which leads to inaccurate small-target position coding information, consequently affecting the prediction of target pixels.

Compared to our SSCA, \textit{SSCA w MH} suffers decreases of 1.15\%, 1.52\%, and 0.73\% in terms of $IoU$, $IoU$, and $F$-$measure$ values on the SIRST-1K dataset. This is because the multi-head strategy complicates the feature mapping space of IR small targets, which is rather unfavorable for extracting information from targets with limited features. Therefore, in SCTransNet, we utilize the single-head attention for IRSTD.

Comparing SSCA and the variant \textit{SSCA w/o SE}, we find that the local spatial embedding can significantly improve the performance of infrared small target detection in the three public datasets.
Visualization maps displayed in Fig.~\ref{fig8} further illustrate the effectiveness of this strategy.
This is due to the ability of local spatial embedding to capture both specific details of the target and potential spatial correlations in the background within the deep layers.
As a result, this approach minimizes instances of missed detections and improves the confidence of the detection process.

\subsubsection{The Complementary Feed-forward Network}
\begin{figure}[t]
    \centering
    \includegraphics[width=8.4cm]{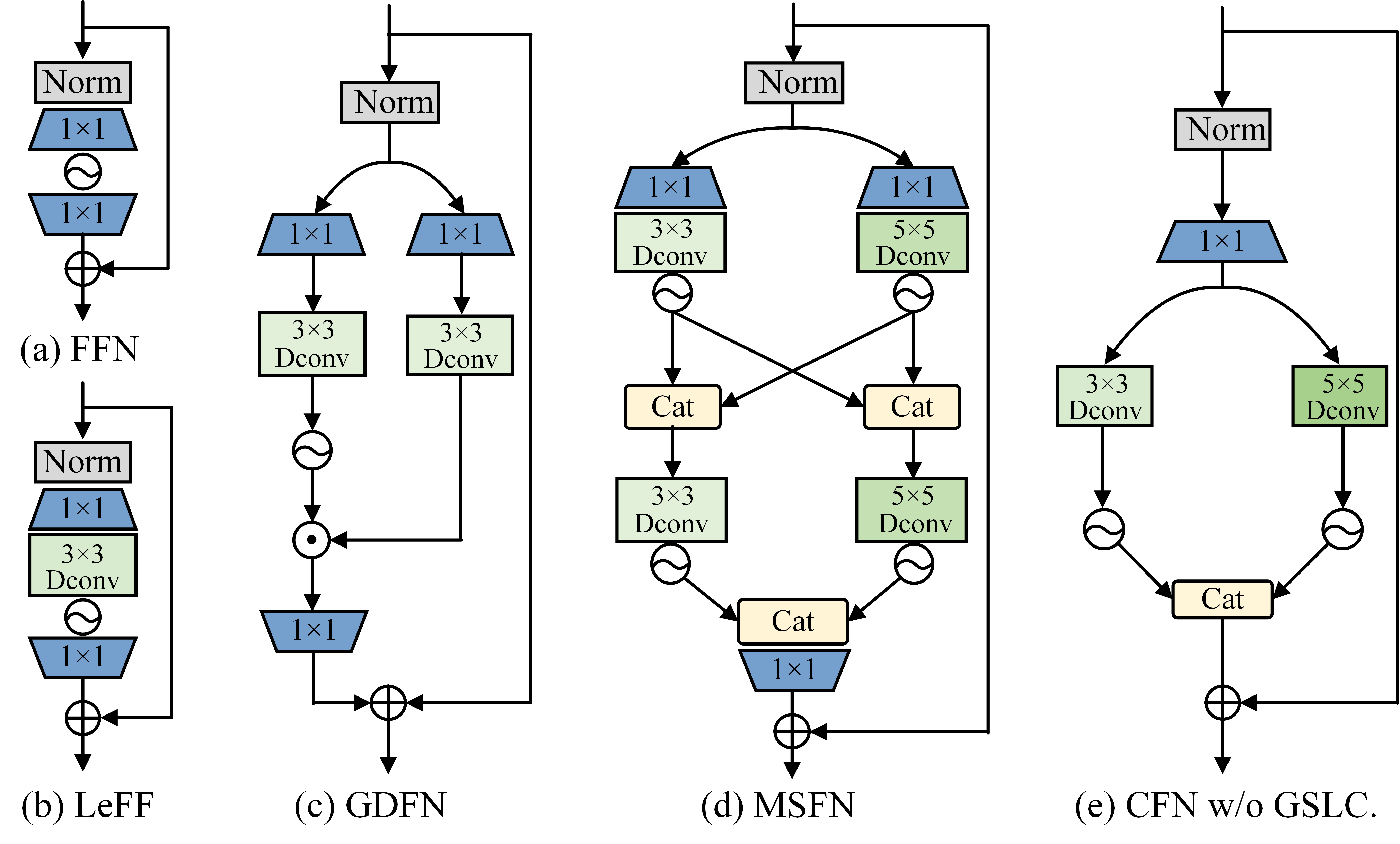}
    \caption{The structure of representative feed-forward networks and CFN w/o GSLC.}
    \label{fig9}
\end{figure}

Feed-forward networks (FFNs) are used to strengthen the information correlation within features and introduce nonlinear radicalization to enrich the feature representation.
In this section, we use five different FFN models based on SCTransNet to compare the proposed CFNs.
As shown in Fig.~\ref{fig9}, we used typical FFN~\cite{14} (ViT for image classification), LeFF~\cite{48} (Uformer for image restoration) embedded in localized space, GDFN~\cite{22} (Restormer for image restoration) based on gated convolution, MSFN~\cite{67} (Sparse transformer for image deraining) based on multi-scale depth-wise convolution, the variant CFN without global spatial and local channel module (\textit{CFN w/o GSLC}), respectively.

\begin{table}[t]
\centering
\caption{$IoU(\%)$/$nIoU(\%)$ values achieved by the representative feed-forward Networks and the variants of CFN on NUAA-SIRST and NUDT-SIRST.}
\label{tab7}
\renewcommand{\arraystretch}{1.3}
\setlength\tabcolsep{1.3mm}
\begin{tabular}{ccccc}    \hline\hline
\multirow{2}{*}{Model}   &\multirow{2}{*}{Params(M)}  &\multirow{2}{*}{Flops(G)} & \multicolumn{2}{c}{Dataset} \\  \cline{4-5} 
                        &           &                 & NUAA-SIRST            & NUDT-SIRST        \\ \hline
FFN~\cite{14}  & \textbf{11.0292}  & \textbf{20.1474}     & 76.87/80.08    & 93.58/93.85   \\ 
LeFF~\cite{48} & 11.1312  & 20.1944     & 76.49/80.21    & 93.92/94.07   \\
GDFN~\cite{22} & 10.1841  & 19.7210    & 75.48/79.32    & 93.40/93.64   \\
MSFN~\cite{67} & 11.7107  & 20.5026    & 77.35/79.89    & 93.88/94.24   \\
CFN w/o GSLC   & 11.1905  & 20.2362    & 76.54/80.56    & 93.95/94.18   \\
CFN            & 11.1905  & 20.2372    & \textbf{77.50/81.08}  & \textbf{94.09/94.38}   \\  
\hline
\end{tabular}
% %\vspace{-3mm}
\end{table}

\begin{figure}[t]
    \centering
    \includegraphics[width=8.6cm]{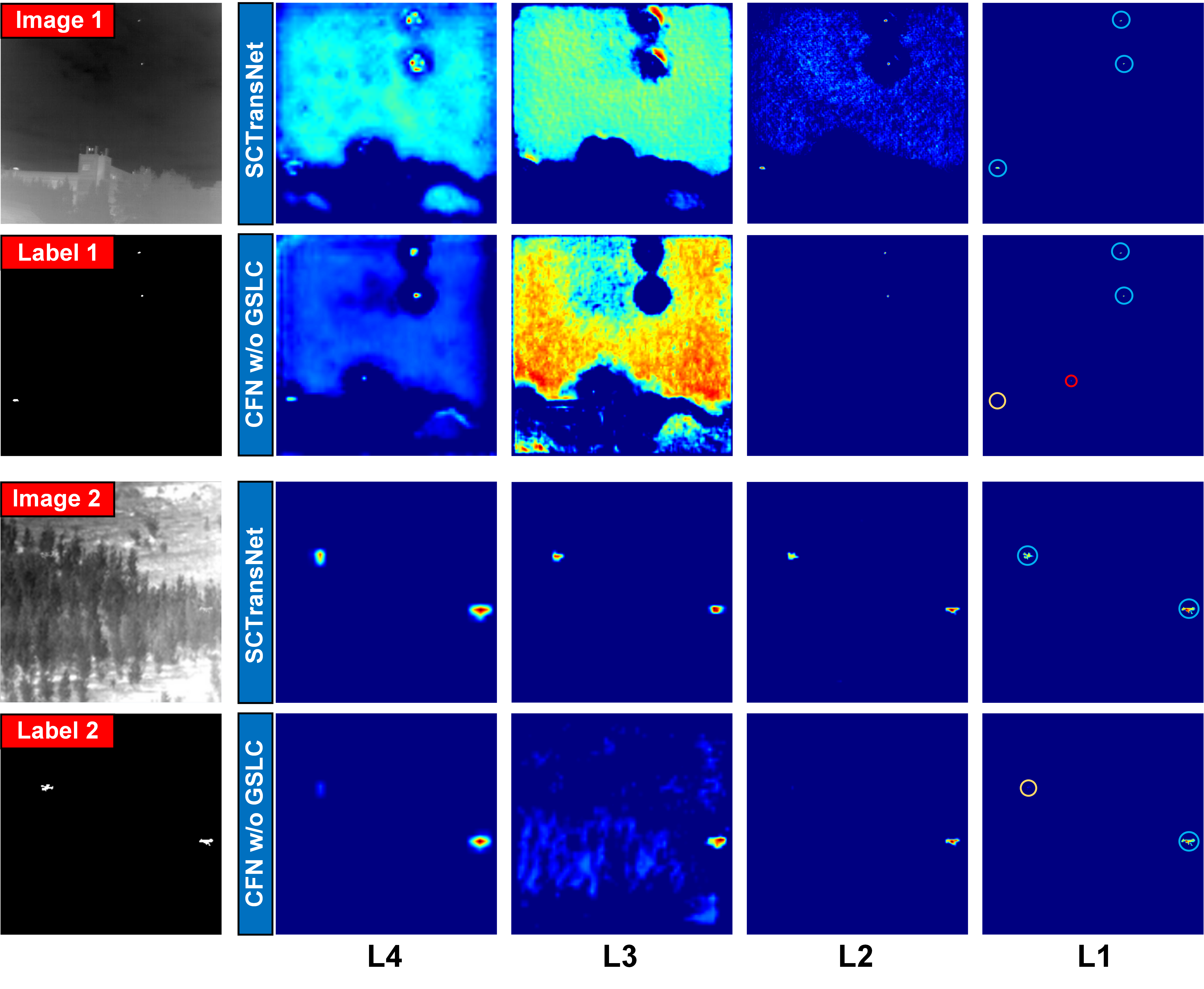}
    % %\vspace{-3mm}
    \caption{Visualization map of SCTransNet and CFN w/o GSLC. The feature maps from the deep layer of CFN w/o GSLC mix targets in the background. It finally results in missed detection in the output layer.}
    % %\vspace{-1mm}
    \label{fig10}
\end{figure}

As shown in Table~\ref{tab7}, LeFF exhibits a slight improvement in metrics over FFN, which indicates that the local spatial information aggregation employed in feed-forward neural networks is effective for IRSTD.
Because gated convolution tends to consider IR small targets as noise and filters them out, this results in the GDFN having a low detection accuracy.
We also find that MSFN outperforms all methods except our CFN, illustrating the superior ability of multi-scale structures to interact with spatial information compared to single-scale structures.
Finally, we observe that the performance of the variant \textit{CFN w/o GSLC} is inferior to that of MSFN. However, when we incorporate the GSLC module, our CFN achieves optimal values of $IoU$ and $nIoU$ on the NUAA and NUDT datasets. Moreover, the network's parameters and computational complexity remain almost unchanged, which demonstrates the validity and utility of the complementary mechanism proposed in this paper for the IRSTD task.
As illustrated in Fig.~\ref{fig10}, with the help of the complementary mechanism, the network allows for more effective enhancement of infrared small targets and suppression of clutter in building and jungle backgrounds, leading to improved target detection accuracy.

\begin{table}[t]
\centering
\caption{$IoU(\%)$/$nIoU(\%)$ values achieved by the different cross-layer feature fusing modules on NUAA-SIRST and NUDT-SIRST.}
\label{tab8}
\renewcommand{\arraystretch}{1.3}
\setlength\tabcolsep{1.4mm}
\begin{tabular}{ccccc}    \hline\hline
\multirow{2}{*}{Model}   &\multirow{2}{*}{Params(M)}  &\multirow{2}{*}{Flops(G)} & \multicolumn{2}{c}{Dataset} \\  \cline{4-5} 
                        &           &                 & NUAA-SIRST            & NUDT-SIRST          \\ \hline
C.ACM~\cite{7}    & 13.0627  & 30.9862    &  75.68/79.52  &   93.92/94.24  \\ 
C.AGPC~\cite{30}  & 11.7581  & 22.9647    & 77.39/79.96   & 94.01/94.22 \\
C.AFFPN~\cite{31} & 11.7171  & 22.7291    & 76.12/79.33   & 93.53/93.69  \\
SCTransNet     & \textbf{11.1905}  & \textbf{20.2372}    & \textbf{77.50/81.08}    & \textbf{94.09/94.38}   \\ \hline
\end{tabular}
\end{table}

\subsubsection{The Impact of CCA Block}
As mentioned in Sec.~\ref{Sec2-A}, cross-layer feature fusion can facilitate the preservation of enhanced target information. 
In this section, we utilize three cross-layer feature fusion structures, namely ACM~\cite{7}, AGPC~\cite{30}, and AFFPN~\cite{31}, derived from different IRSTD methods, to replace the CCA module employed in SCTransNet. This substitution yields the variation structures, namely C.ACM, C.AGPC, and C.AFFPN, respectively.
As shown in Table~\ref{tab8}, the results illustrate that our SCTransNet obtains the highest IoU and nIoU values on the NUAA and NUDT datasets with the lowest model parameters and computational complexity. This illustrates the effectiveness of the CCA we utilized.

\subsection{Core Hyper-parameter Analysis}
We utilize the depth of the RBs, the number of SCTBs, the channel expansion factor of CFNs, and the base width of the model to validate the hyper-parameters of SCTransNet.
As shown in Table~\ref{tab9}, the numbers ``0”, ``1”, ``2”, and ``3” indicate the embedding depth of the RBs. We observe that as the \replaced{residual block}{ResBlock} depth increases, there is a slight increase in both the number of parameters and flops, and the performance of IRSTD shows significant improvement.
This improvement can be attributed to the residual connection facilitating gradient propagation and mitigating feature degradation. Therefore, our SCTransNet uses four \replaced{residial blocks}{ResBlocks} for information encoding.
Table~\ref{tab10} illustrates the results of the hyper-parameter study of the number of SCTBs, the channel expansion factor of CFNs, and the basic width of the model. 
It is evident that as the number of SCTB modules increases, the model's performance steadily improves, reaffirming the effectiveness of the SCTB model.
We observe that while the performance with 6 SCTBs is slightly better than with 4 SCTBs, it incurs excessive computational complexity. When the channel expansion factor $\eta$ = 2.66, the model can get the best performance. Additionally, we also noticed that setting the base width of the model W=48 results in a slight degradation in performance compared with W=32, which can be attributed to the excessive model parameters reducing the algorithm's generalization ability.
Therefore, in our proposed SCTransNet, the number of SCTBs, the channel expansion factor of CFNs, and the base width of the model are set to 4, 2.66, and 32, respectively.

\begin{table}[t]
\centering
\caption{Hyper-parameter study of the RBs in average $IoU(\%)$, $nIoU(\%)$, $F$-$measure(\%)$ on NUAA-SIRST, NUDT-SIRST, and IRSTD-1K.}
\label{tab9}
\renewcommand{\arraystretch}{1.3}
\setlength\tabcolsep{1.5mm}
\begin{tabular}{ccccccccc}  \hline \hline
1 & 2 & 3 & 4 & IoU   & nIoU  & F-measure & Params(M) & Flops(G) \\ \hline
\ding{51}& \ding{55} & \ding{55} & \ding{55} & 82.29 & 85.77 & 90.26     & 20.0212   & 11.1462  \\ \hline
\ding{51} & \ding{55} & \ding{55} & \ding{55} & 82.33 & 85.89 & 90.31     & 20.0967   & 11.1484  \\ \hline
\ding{51} & \ding{51} & \ding{55} & \ding{55} & 82.49 & 86.11 & 90.40     & 20.1680   & 11.1569  \\ \hline
\ding{51} & \ding{51} & \ding{51} & \ding{55} & 82.95 & 86.27 & 90.68     & 20.2372   & 11.1905  \\ \hline
\ding{51} & \ding{51} & \ding{51} & \ding{51} & 83.43 & 86.86 & 90.96     & 20.2372   & 11.1905  \\ \hline
\end{tabular}
\end{table}

\begin{table}[t]
\centering
\caption{Hyper-parameter study of the number of SCTBs, the channel expansion factor of CFN, and the basic width of the model in average $IoU(\%)$, $nIoU(\%)$, $F$-$measure(\%)$ on NUAA-SIRST, NUDT-SIRST, and IRSTD-1K.}
\label{tab10}
\renewcommand{\arraystretch}{1.3}
\setlength\tabcolsep{1.5mm}
\begin{tabular}{cccccc} \hline \hline
Hyper-param & IoU   & nIoU  & F-measure & Params(M) & Flops(G) \\ \hline
\multicolumn{6}{c}{The number of SCTBs}                    \\ \hline
N = 1       & 82.33 & 85.86 & 90.28     & 17.7408   & 6.3295   \\ \hline
N = 2       & 82.53 & 86.05 & 90.43     & 18.5729   & 7.9498   \\ \hline
N = 3       & 82.97 & 86.46 & 90.58     & 19.4051   & 9.5702   \\ \hline
\textbf{N = 4}       & 83.43 & 86.86 & 90.96     & 20.2372   & 11.1905  \\ \hline
N = 5       & 83.40 & 86.84 & 90.95     & 21.0694   & 12.8108  \\ \hline
N = 6       & 83.45 & 86.86 & 90.97     & 21.9015   & 14.4312  \\ \hline
\multicolumn{6}{c}{The channel expansion factor of CFNs}    \\ \hline
$\eta$ = 1.33    & 82.80 & 86.18 & 90.59     & 19.2457   & 9.2539   \\ \hline
$\eta$ = 2.00    & 82.75 & 86.32 & 90.56     & 19.7474   & 10.2338  \\ \hline
\textbf{$\eta$ = 2.66}   & 83.43 & 86.86 & 90.96     & 20.2372   & 11.1905  \\ \hline
$\eta$ = 3.00    & 83.24 & 86.69 & 90.84     & 20.4938   & 11.6917  \\ \hline
$\eta$ = 3.99    & 83.10 & 86.60 & 90.77     & 21.2306   & 13.1307  \\ \hline
\multicolumn{6}{c}{The basic width of the model}           \\ \hline
W = 8       & 77.52 & 80.55 & 87.33     & 1.3321    & 0.7468   \\ \hline
W = 16      & 81.02 & 84.50 & 89.51     & 5.1488    & 2.8609   \\ \hline
\textbf{W = 32}      & 83.43 & 86.86 & 90.96     & 20.2372   & 11.1905  \\ \hline
W = 48      & 82.95 & 86.48 & 90.60     & 45.2687   & 24.994   \\ \hline
\end{tabular}
\end{table}

\subsection{Robustness of SCTransNet}
\begin{figure}[t]
    \centering
    \includegraphics[width=8.4cm]{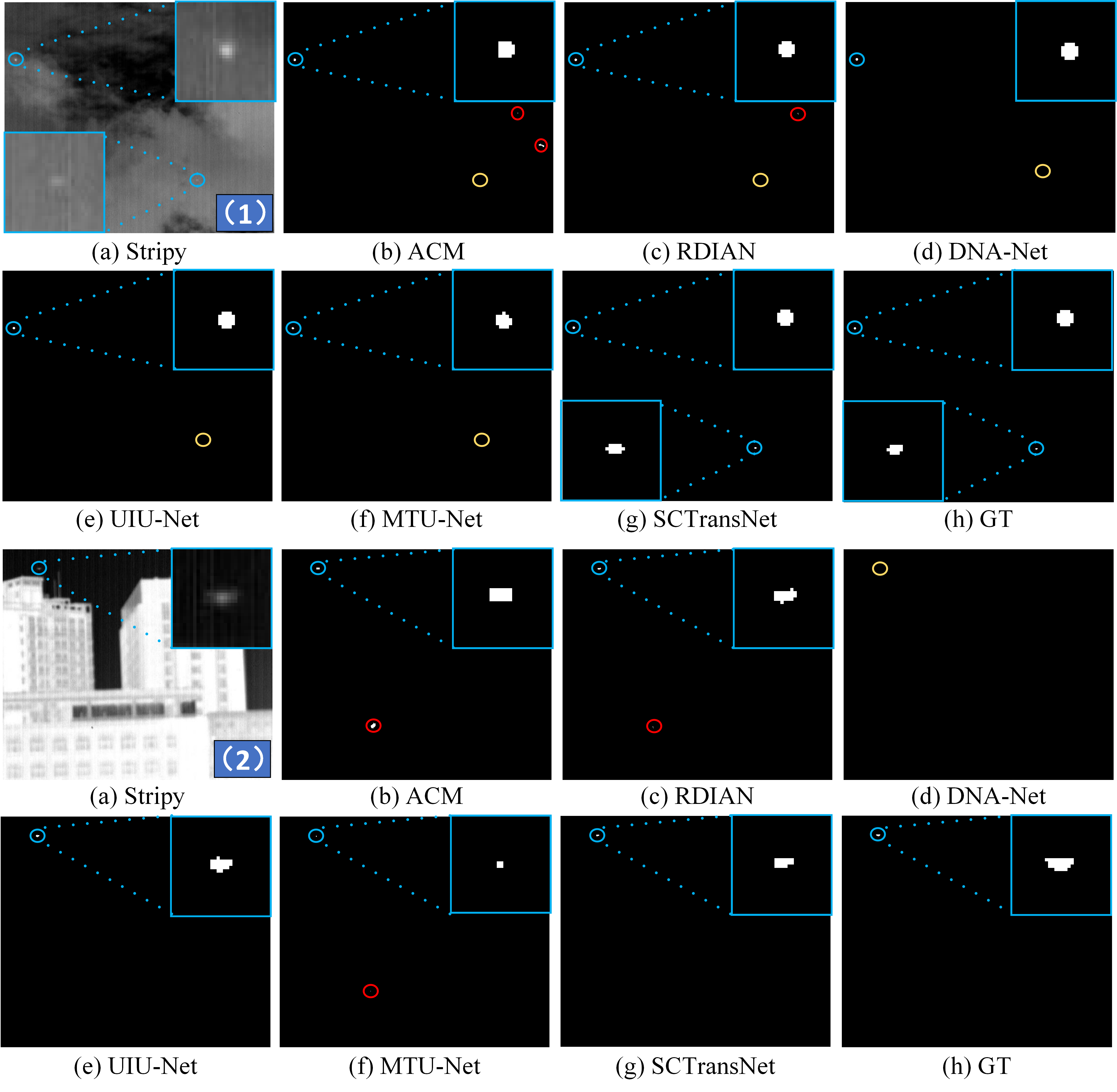}
    \caption{Visual results obtained by different IRSTD methods on the real stripy IR images. Circles in blue, yellow, and red represent correctly detected targets, miss detections, and false alarms, respectively.}
    \label{fig11}
\end{figure}

In an actual IR detection system, the non-uniform response of the focal plane array (FPN) can cause stripe noise in IR images~\cite{80}. This presents a challenge to the noise immunity and generalization ability of the IRSTD methods.
Fig.~\ref{fig11} gives the visual effect of the IR image with real stripe noise on various detection methods. It is evident that the noise destroys the local neighborhood information of the targets.
In Fig.~\ref{fig11}(1), only our SCTransNet accurately detects two targets, while the other methods exhibit missed detections and false alarms.
In Fig.~\ref{fig11}(2), there is also a piece of blind element in the striped image, which interferes with the semantics understanding of the building. As a result, the ACM, RDIAN, and MTU-Net generate false alarms around the blind element.
The ability to explicitly establish full-level contextual information about the target and the background is what makes our approach more robust.

\section{CONCLUSION}
In this paper, we presented a Spatial-channel Cross Transformer Network (SCTransNet) for IR small target detection. 
Our SCTransNet utilizes spatial-channel cross transformer blocks to establish associations between encoder and decoder features to predict the context difference of targets and backgrounds in deeper network layers.
We introduced a spatial-embedded single-head channel-cross attention module, which establishes the semantic relevance between targets and backgrounds by interacting local spatial features with global full-level channel information.
We also devised a complementary feed-forward network, which employs a multi-scale strategy and crosses spatial-channel information to enhance feature differences between the target and background, thereby facilitating effective mapping of IR images to the segmentation space.
Our comprehensive evaluation of the method on three public datasets shows the effectiveness and superiority of the proposed technique.

\bibliographystyle{IEEEtran}
\small\bibliography{IEEEabrv, reference}
\vspace{11pt}

 \vspace{-8mm}
 
 \begin{IEEEbiography}[{\includegraphics[width=1in, height=1.4in, clip, keepaspectratio]{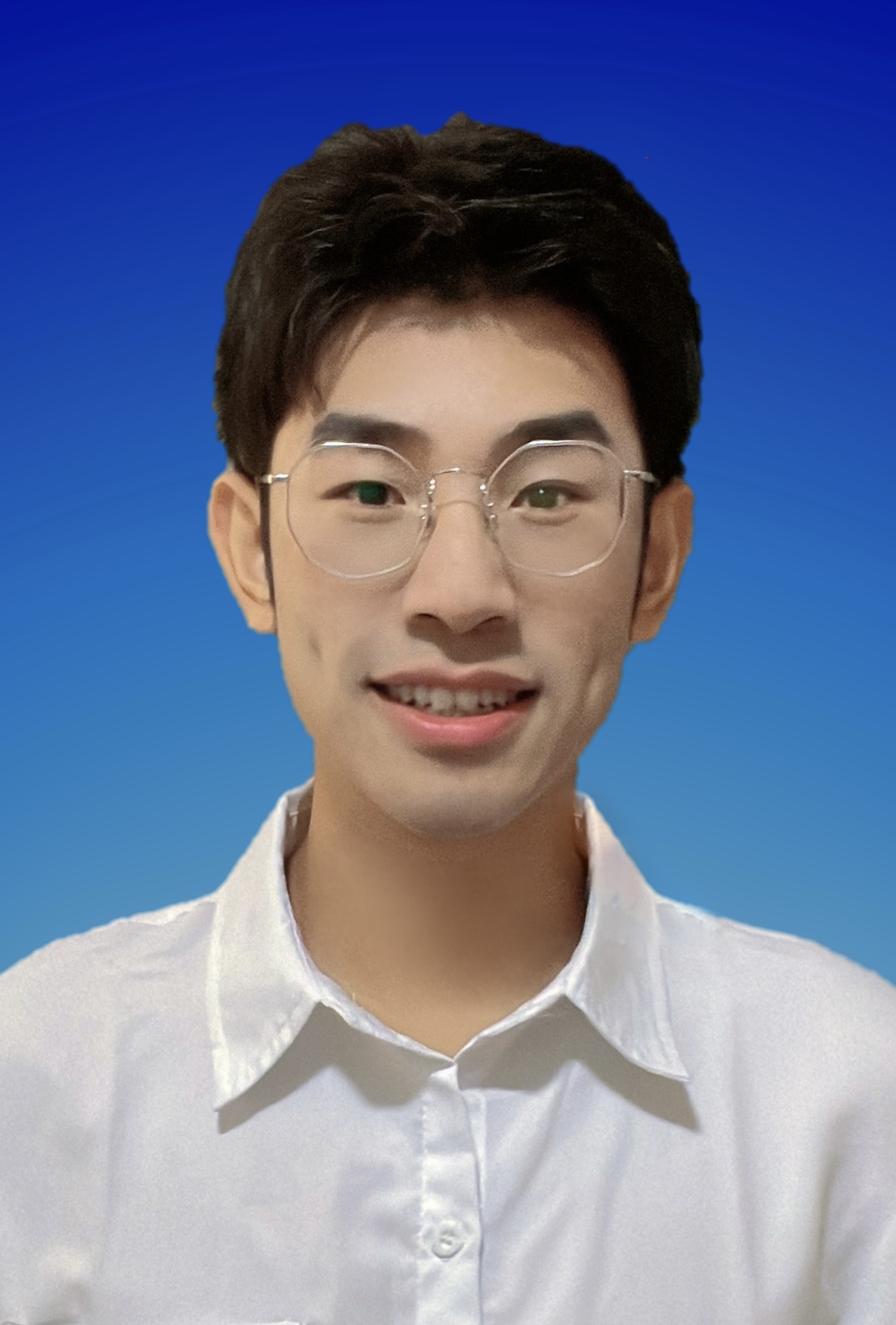}}]{Shuai Yuan}
 received the B.S. degree from Xi'an Technological University, Xi'an, China, in 2019. He is currently pursuing a Ph.D. degree at Xidian University, Xi’an, China. He is currently studying at the University of Melbourne as a visiting student, working closely with Dr. Naveed Akhtar. His research interests include infrared image understanding, remote sensing, and deep learning.
 \end{IEEEbiography}
 \vspace{-5mm}

 \begin{IEEEbiography}[{\includegraphics[width=1in, height=1.4in, clip, keepaspectratio]{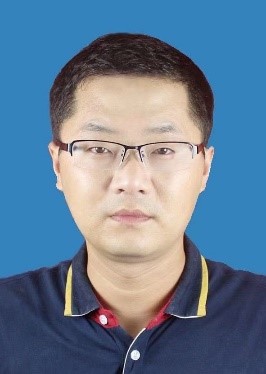}}]{Hanlin Qin}
received the B.S and Ph.D. degrees from Xidian University, Xi'an, China, in 2004 and 2010. He is currently a full professor at the School of Optoelectronic Engineering, Xidian University. He authored or co-authored more than 100 scientific articles. His research interests include electro-optical cognition, advanced intelligent computing, and autonomous collaboration.
\end{IEEEbiography}
 \vspace{-4mm}

 \begin{IEEEbiography}[{\includegraphics[width=1.5in, height=1.5in, clip, keepaspectratio]{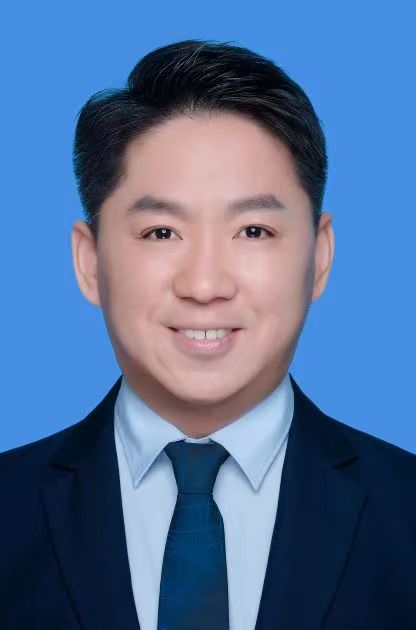}}]{Xiang Yan} received the B.S and Ph.D. degrees from Xidian University, Xi'an, China, in 2012 and 2018. He was a visiting Ph.D. Student with the School of Computer Science and Software Engineering, Australia, from 2016 to 2018, working closely with Prof. Ajmal Mian. He is currently an associate professor at Xidian University, Xi'an, China. His current research interests include image processing, computer vision and deep learning.
\end{IEEEbiography}
 \vspace{-4mm}

 \begin{IEEEbiography}[{\includegraphics[width=1in, height=1.2in, clip, keepaspectratio]{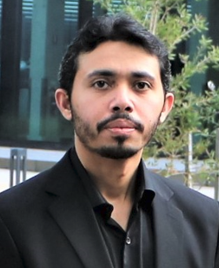}}]{Naveed Akhtar} is a Senior Lecturer  at the University of Melbourne. He received his PhD in Computer Science from the University of Western Australia and Master degree from Hochschule Bonn-Rhein-Sieg, Germany. He is a recipient of the Discovery Early Career Researcher Award from the Australian Research Council. He is a Universal Scientific Education and Research Network Laureate in Formal Sciences. He was a finalist of the Western Australia's Early Career Scientist of the Year 2021.  He is an ACM Distinguished Speaker and serves as an Associate Editor of IEEE Trans. Neural Networks and Learning Systems.
 \end{IEEEbiography}
 \vspace{-5mm}

\begin{IEEEbiography}[{\includegraphics[width=1in, height=1.2in, clip, keepaspectratio]{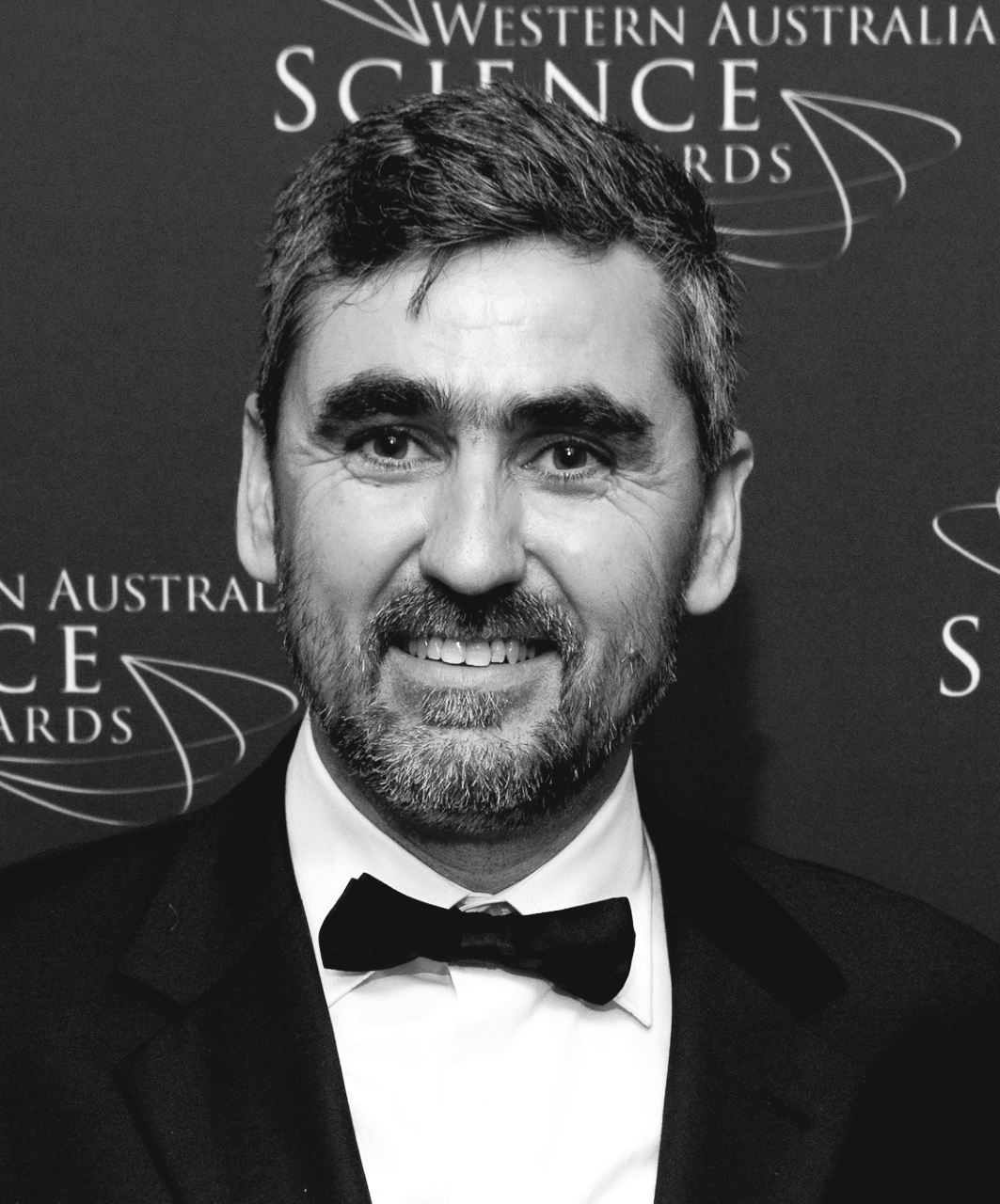}}]{Ajmal Mian} is a Professor of Computer Science at The University of Western Australia. He is the recipient of three esteemed national fellowships from the Australian Research Council (ARC) including the recent Future Fellowship Award 2022. He is a Fellow of the International Association for Pattern Recognition and recipient of several awards including the West Australian Early Career Scientist of the Year Award 2012, the HBF Mid-Career Scientist of the Year Award 2022, Excellence in Research Supervision Award, EH Thompson Award, ASPIRE Professional Development Award, Vice-chancellors Mid-career Research Award, Outstanding Young Investigator Award, and the Australasian Distinguished Doctoral Dissertation Award. Ajmal Mian has secured research funding from the ARC, NHMRC, DARPA, and the Australian Department of Defence. He has served as a Senior Editor for IEEE Transactions on Neural Networks \& Learning Systems and Associate Editor for IEEE Transactions on Image Processing and the Pattern Recognition journal. His research interests include computer vision, machine learning, remote sensing, and 3D point cloud analysis.
 \end{IEEEbiography}

\vfill

\end{document}